%% file: main.tex
\PassOptionsToPackage{table}{xcolor}

\documentclass[11pt,letterpaper,logo]{thuair}

\graphicspath{{./}}

\usepackage[numbers,sort&compress]{natbib}

\usepackage[section]{placeins}

\usepackage{wrapfig}


\input{tex/packages_macros}
\title{\sys$\zeta$: An Efficient Closed-Loop Embodied Harness for Self-Evolving Physical Intelligence}

\makeatletter
\def\@fnsymbol#1{\ensuremath{\ifcase#1\or *\or \dagger\or \ddagger\or
    \mathsection\or \mathparagraph\or \|\or **\or \dagger\dagger
\or \ddagger\ddagger \else\@ctrerr\fi}}
\makeatother

\author{%
  Xin Ding$^{*}$, Liang Mi$^{*}$, Mingzhe Huang$^{*}$, Zixuan Wang$^{*}$, Chao Zhang$^{*}$, Zixu Hao$^{*}$, Fu Chen, Xiangyu Li, Yikai Zheng, Yaoyu Guo, Weijun Wang, Kun Li, Hao Wu$^{\dagger,\ddagger}$, Yunxin Liu$^{\dagger}$, Ting Cao$^{\dagger}$ \\
  \textbf{Institute for AI Industry Research (AIR), Tsinghua University; \quad Z-Trans AI}\\
  \textbf{$^{*}$Equal contribution \quad $^{\dagger}$Corresponding author \quad $^{\ddagger}$Work done during a visit to AIR, Tsinghua}\\
 \textbf{Technical lead: Xin Ding, Liang Mi}\\
 \textbf{Project lead: Ting Cao (tingcao@mail.tsinghua.edu.cn)}\\
  \textbf{Project Page:} \url{https://air-embodied-brain.github.io/zetta}
  \vspace{-12pt}
}

\begin{document}
\newlength{\thuairnormalheadsep}
\setlength{\thuairnormalheadsep}{\headsep}
\setlength{\headsep}{16pt}

\input{tex/abstract}

\let\thuairabscontent\abscontent
\renewcommand{\abscontent}{}
\maketitle
\let\abscontent\thuairabscontent

\enlargethispage{30\baselineskip}

\begin{figure}[!h]
  \centering
  \includegraphics[width=0.8\linewidth]{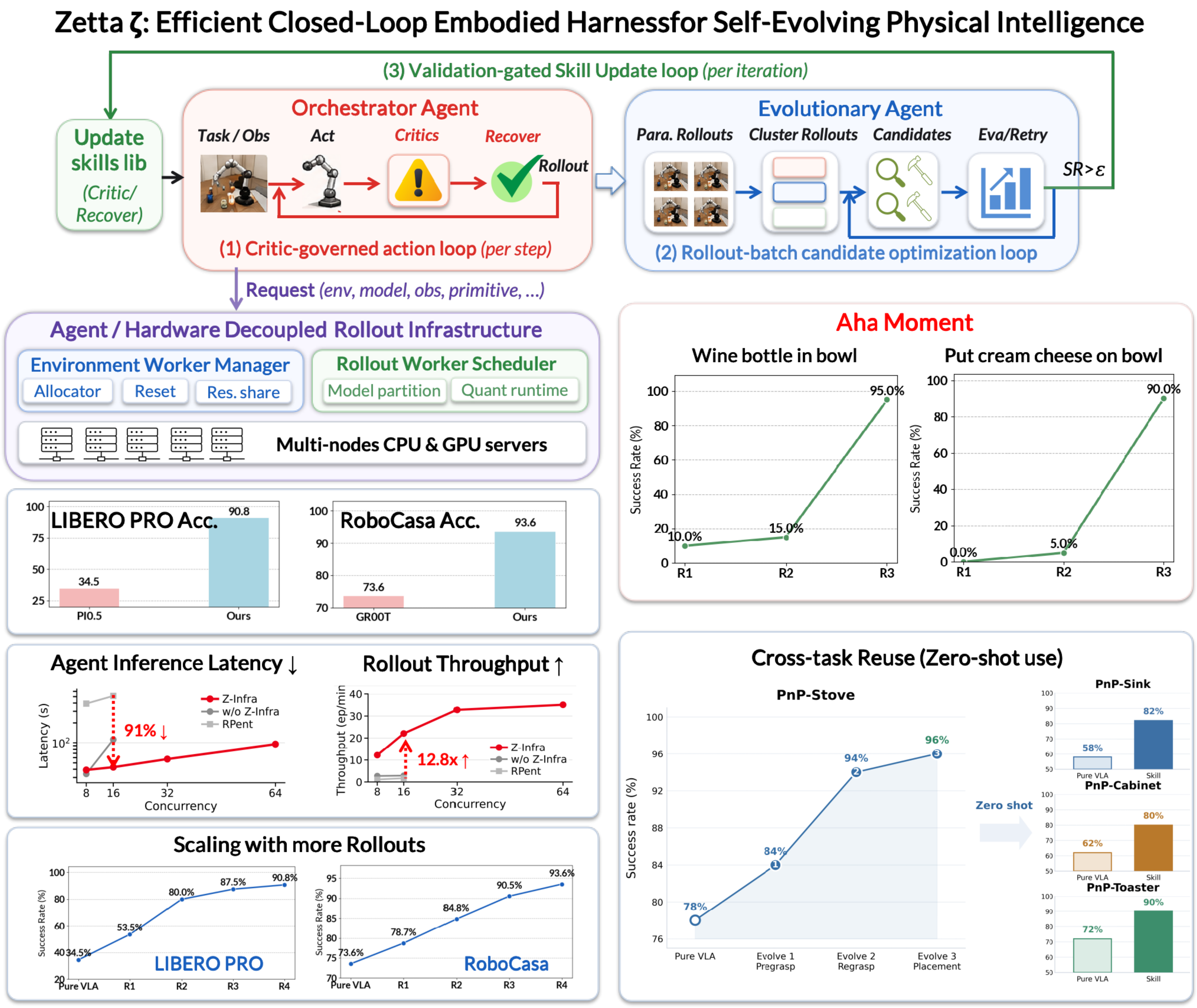}
  \captionsetup{font=footnotesize}
  \caption{\textbf{\sys closes the loop for embodied self-evolution. Frequent runtime critics trigger recoveries during execution, while verified failures are distilled into reusable critic and recovery skills across rollouts. A hardware-decoupled rollout layer scales this process across heterogeneous environments, models, CPUs, and GPUs. \sys enables sustained same-task improvement, zero-shot skill transfer, robotic "Aha moment", and accelerated agent execution.}}
  \label{fig:teaser}
\end{figure}

\abscontent
\newpage

\begingroup
\small
\linespread{0.7}\selectfont
\setcounter{tocdepth}{2}
\hypersetup{linkcolor=black}
\tableofcontents
\endgroup
\newpage

\input{tex/introduction}

\input{tex/method}
\input{tex/experiment}
\input{tex/related}
\input{tex/conclusion}

\section*{Acknowledgment}
We thank Fucheng Jia, Mingju Wang, An Pan, Zexu Wang, Zijian Wang, Shuhao Wu, Wenhui Gu, Jiawei He, Yi Tao, Wei Sun for their contributions on engineering and real robot demos.

\appendix
\input{tex/appendix}


\input{main.bbl}
\end{document}

%% file: tex/packages_macros.tex
\usepackage{xspace}
\usepackage{graphicx}
\usepackage{subcaption}
\usepackage{multirow}
\usepackage{makecell}
\usepackage{amsmath}
\usepackage{amssymb}
\usepackage{enumitem}

\usepackage[ruled,linesnumbered,noend]{algorithm2e}
\usepackage{hyperref}
\usepackage{listings}
\lstdefinestyle{rolloutflow}{
  basicstyle=\ttfamily\small,
  backgroundcolor=\color{gray!10},
  frame=single,
  framerule=0.4pt,
  rulecolor=\color{gray!50},
  breaklines=true,
  columns=fullflexible,
  keepspaces=true,
  commentstyle=\color[RGB]{0,128,0},
  morecomment=[l]{\#}
}

\newcommand{\sys}{Zetta\xspace}

%% file: tex/abstract.tex
\begin{abstract}
Embodied agents are increasingly used to close the gap left by end-to-end policy models. Yet the agentic path has not realized closed-loop learning in physical execution: existing harnesses remain largely open-loop, following fixed skills during rollout and reflecting only after an episode completes. Such post-hoc reflection cannot govern execution as it unfolds, because physical interaction requires decisions to track rapidly changing robot-environment states at a frequency beyond today’s large agentic models. We present \sys, a closed-loop embodied harness that evolves code-based runtime critics and recovery skills online while keeping the base policy frozen. Through three timescale-separated loops, Zetta provides action-frequency governance, rollout-level critic-recovery proposal, and validation-gated skill updates. Together with Z-Infra, a rollout infrastructure decoupling agent logic from heterogeneous execution resources, Zetta achieves state-of-the-art success on LIBERO-Pro and RoboCasa under our current rollout budget, reaching 90.8\% and 93.6\%, with a $11.1\times$ inference speedup; success continues to scale with self-exploration experience; learned skills transfer zero-shot, and clear robotic ``Aha Moments'' emerge. These results show that closed-loop harness self-evolution opens a scaling path for reliable physical intelligence.
\end{abstract}

%% file: tex/introduction.tex

\section{Introduction}

Physical intelligence is advancing along two complementary paths for effective scaling. The first scales end-to-end policy models, including vision-language-action models (VLAs) and world-action models (WAMs)~\cite{brohan2022rt1,brohan2023rt2,kim2024openvla,black2024pi0,black2025pi05,bjorck2025groot,li2024cogact,bu2025univla,pertsch2025fast,dreamzero2026,cosmospolicy2026,fastwam2026,yu2026wall,team2026xiaomi,lingbot_va,lingbot_vla,qwen_vla,qwen_robotmanip,kim2026rldx,jiang2025galaxea,zheng2025xvla,nvidia2026cosmos3}, trained on large-scale demonstration corpora such as DROID and Open X-Embodiment~\cite{khazatsky2024droid,oneill2023openx}.
However, embodied data remain scarce, and models trained on limited data distributions are brittle under the changing dynamics of real-world deployment~\cite{belkhale2023dataquality,ross2011reduction,nagabandi2018metarl,mandlekar2021whatmatters,simchowitz2025pitfalls,xie2023decomposinggap}; the gap from demonstration to reliable real-world execution therefore remains open.

The second path uses large language models as embodied agents to orchestrate policy models, code, tools, and control primitives~\cite{capx2026,aspire2026,xiao2026enpire,anthropic2026clauderobotics,zhang2026harnessvla}, with the aim of learning from self-exploration in environments. Though these systems alleviate limitations of end-to-end policies, existing embodied agent harnesses do not actually realize this learning. Existing embodied agent harnesses therefore remain largely open-loop: once execution begins, the agent does not continuously condition its decisions on the evolving robot-environment state, but instead follows fixed skills or preplanned trajectories and reflects only after the episode is completed. \emph{This work enables a closed-loop embodied agent harness, learning from self-exploration and demonstrating self-evolution: as rollout experience increases, the success rate continues to improve.}

Realizing closed-loop execution for embodied agents is fundamentally bottlenecked by the high-frequency interaction between robots and changing environments. Physical tasks require decisions to be coupled to the current robot and environment state, often changing within a millisecond-level latency budget~\cite{li2025holdmybeer,lee2022realtimempc}. Large agentic models cannot make decisions at this frequency in real-world systems~\cite{zhang2024hirt,yan2026actingwhileunderstanding,tan2024fouriercontroller}. Existing methods therefore perform reflection mainly at the episode or trajectory level, which can diagnose a completed failure but cannot govern the physical execution while it unfolds~\cite{shinn2023reflexion,liu2023reflect,embodiskill2026,hong2026reflectiveplanning,feng2026procvlm}. Such post-hoc reflection has inherent limitations: agents cannot online test alternative actions to verify whether a reflection is correct, credit assignment over an entire trajectory is difficult, and retrospective analysis often lacks access to the precise state at the moment of failure. As a result, the experiences summarized after execution are difficult to reuse and provide limited support for effective learning. The experience adapts poorly to the changing real-world dynamics~\cite{weng2025temporalactionselection,julian2020neverstoplearning,chen2023adaptonthego}.

To address this challenge, the key idea of this work is to introduce \textbf{code-based critics that evolve online for closed-loop execution}, governing physical execution and triggering agentic intervention according to the changing dynamics of actions and environments. Building on this idea, we present \sys, a closed-loop embodied harness that self-evolves to scale physical intelligence (Figure~\ref{fig:teaser}).  The base policy models are frozen while evolving a harness $H=\{C,R,T\}$ of runtime critics, corresponding recovery skills from the rollouts. Specifically, we propose three different time-scale loops to enable self-evolution. First, a \emph{Critic-Governed Action Loop} executes learned critics at action frequency and invokes the corresponding recovery skills when necessary. Second, a \emph{Rollout-Batch Candidate Optimization Loop} clusters and diagnoses failures from each iteration, then proposes candidate critics and recoveries. Because skills and code provide no gradients, we draw on SkillOpt and EmbodiSkill~\cite{yang2026skillopt,embodiskill2026} to construct an SGD-like optimization process that makes bounded, stable updates in code space. Third, a \emph{Validation-Gated Skill Update Loop} admits only critic and recovery candidates that improve success rate and generalize across rollouts, and then adds them to the skill memory. The first loop enables closed-loop execution, while the second and third loops realize self-evolution.

The closed-loop enables the harness learning from self-exploration. This learning process also makes rollout throughput the rate limiter of intelligence scaling: the environment is the source of learning data, so faster rollouts produce faster evolution. We therefore propose and realize Z-Infra, to our knowledge the first rollout infrastructure designed specifically for self-evolving embodied agents. It decouples agent logic from heterogeneous execution resources through independent environment and model worker pools, batched inference, model partitioning, and asynchronous scheduling. This design scales the same agent across machines, accelerators, models, and environments without binding its logic to a particular hardware configuration.

Our main results are:
\begin{itemize}
\item Under our current rollout budget, \sys achieves state-of-the-art task success, substantially improving over current policy models and embodied-agent baselines: 90.8\% on LIBERO-Pro and 93.6\% on RoboCasa. Since performance continues to improve across evolution rounds, further rollout experience is expected to yield additional gains.
\item Success scales with evolution iterations: success increases from 34.5\% to 90.8\% on
LIBERO-Pro and from 73.6\% to 93.6\% on RoboCasa within several evolution iterations. 
\item The learned skills demonstrate zero-shot transfer capability to similar tasks. Take the PnP-Stove task in RoboCasa as an example, the pregrasp, regrasp, and stable-placement skills learned during the evolution of
transfer zero-shot to three related PnP tasks without target-specific
optimization: success improves from 58\% to 82\% on PnP-Sink, from 62\% to
80\% on PnP-Cabinet, and from 72\% to 90\% on PnP-Toaster.
\item We observe clear ``aha moments'' in which success remains low during
early evolution rounds but rises abruptly once the agent discovers the key
critic--recovery mechanism: for example, from 15\% to 95\% on Wine Bottle in Bowl and from
5\% to 90\% on Put Cream Cheese on Bowl.
  \item The \sys inference latency decreases by 91\% compared to RPent, i.e., an $11.1\times$ speedup. 
  \item Z-Infra increases valid rollout throughput from 1.7 to 35.1 episodes/min, a 20.6$\times$ improvement, directly accelerating the harness evolution loop.
\end{itemize}

Our contributions are: 
\begin{itemize}
  \item The first to realize a self-evolving embodied agent through learnable code-based critics and three coordinated evolution loops operating at action, rollout-batch, and iteration timescales.
  \item The first infrastructure designed to support embodied-agent self-evolution, decoupling agent logic from heterogeneous hardware resources for scalable rollout generation and execution.
  \item We demonstrate the scaling capability of the embodied harness: with increasing rollout and evolution iterations, it progressively improves task success to SOTA among existing models or embodied agents.
\end{itemize}

%% file: tex/method.tex
\section{Enabling Deployment-Time Evolution of Embodied Agents}
\subsection{Problem Formulation: Dual-Agent Governance and Evolutionary Architecture}
\label{sec:problem}
We formalize the long-horizon robotic manipulation task as an authority-constrained governance and evolution process. The framework incorporates an online \textbf{Orchestrator Agent} for adjudication and offline \textbf{Evolutionary Agents} for optimization, enabling closed-loop capability evolution without altering the underlying policy parameters.

\subsubsection{Fixed Runtime Components}
The execution is driven by two core entities whose internal parameters and logic remain invariant throughout the evolution:
\begin{enumerate}
    \item \textbf{Action Policy ($\pi$)}: Given observation $s_t$ and goal $g$, it generates low-level actions $a_t = \pi(s_t, g; \theta)$, where parameters $\theta$ satisfy the constraint $\nabla \theta = 0$.
    \item \textbf{Orchestrator Agent ($\mathcal{A}_{orch}$)}: Acting as a fixed multimodal reasoning operator \cite{anthropic2025claudesonnet45,openai2026gpt55,openai2026gpt56}, $\mathcal{A}_{orch}$ functions as a high-level commander responsible for auditing real-time evidence and approving mode transitions. Its decision logic remains constant during evolution.
\end{enumerate}

\subsubsection{Evolvable Harness ($\mathcal{H}$)}
The Harness $\mathcal{H}$ is the target of the evolutionary process, providing $\mathcal{A}_{orch}$ with means to perceive and intervene in the environment. We define it as:
\begin{equation}
    \mathcal{H} = \{C, R, \mathcal{T}\}
\end{equation}
\begin{itemize}
    \item \textbf{Runtime Critic ($C$)}: A set of \textbf{high-frequency} monitoring functions that persistently scan the trajectory $\tau_{0:t}$ to generate a \textbf{structured proposal} $P_t$:
    \begin{equation}
        P_t = C(\tau_{0:t}) = \langle e_t, \hat{\sigma}_t \rangle
    \end{equation}
    where $e_t$ represents auditable evidence of failure (e.g., collisions, stalled progress) and $\hat{\sigma}_t$ is the suggested execution mode.
    \item \textbf{Recovery Playbook ($R$)}: A structured library of strategies mapped to specific causal failure mechanisms, providing actionable options for adjudication.
    \item \textbf{Heterogeneous Toolset ($\mathcal{T}$)}: A collection of executable tools and operators (e.g., planners, grasp detectors, recovery modules)\cite{zucker2013chomp, sundermeyer2021contact,kirillov2023segment} that can be generated, instantiated, selected, and refined during evolution. The toolset evolves to satisfy the monitoring requirements of $C$ and the recovery logic of $R$, enabling the harness to acquire and adapt its execution capabilities rather than merely tuning parameters of predefined tools.
\end{itemize}

\subsubsection{Authority Hierarchy: Online Adjudication Logic}
During execution, the final operational mode $\sigma_t$ ($\sigma_t = 0$ for VLA or WAM, $\sigma_t > 0$ for specialized tools) is determined by the adjudication function:
\begin{equation}
    \sigma_t = \mathcal{A}_{orch}(P_t, R, \mathcal{T}, \mathcal{K})
\end{equation}
where $P_t$ is the real-time proposal from $C$, and $\mathcal{K}$ represents the \textbf{task knowledge context}, including pre-defined milestones, success criteria, and environmental constraints. This protocol enforces \textbf{evidence-driven decision-making}: although $C$ operates at a high frequency, an intervention is only permitted if the evidence $e_t$ is validated and accepted by $\mathcal{A}_{orch}$.

\subsubsection{Offline Optimization: Evolutionary Agents}
To optimize $\mathcal{H}$, we introduce \textbf{Evolutionary Agents} $\mathcal{A}_{evo}$ as the driving force for offline refinement. $\mathcal{A}_{evo}$ iteratively improves the harness components by analyzing the failed rollout data $\mathcal{D}_{fail}$:
\begin{equation}
    \mathcal{H}^{(k+1)} \leftarrow \mathcal{A}_{evo}(\mathcal{D}_{fail}^{(k)}, \mathcal{H}^{(k)})
\end{equation}
$\mathcal{A}_{evo}$ extracts causal mechanisms from failure evidence, generates repair patches, and abstracts local experiences into versioned harness. It modifies the runtime behavior of $\mathcal{A}_{orch}$ by restructuring its available sensing ($C$) and action ($R, \mathcal{T}$) space.

\paragraph{Optimization Objective}
The ultimate goal of Zetta is to find the optimal configuration $\mathcal{H}^*$ that maximizes the expected task success rate $J$ while keeping $\pi_{VLA}$ and $\mathcal{A}_{orch}$ invariant:
\begin{equation}
    \max_{\mathcal{H}} J(\mathcal{H}) = \mathbb{E}_{g, s_0 \sim \mathcal{D}} \left[ \text{Success}(\tau) \mid \pi, \mathcal{A}_{orch}, \mathcal{H} \right]
\end{equation}

\begin{figure}[t]
  \centering
  \includegraphics[width=\linewidth]{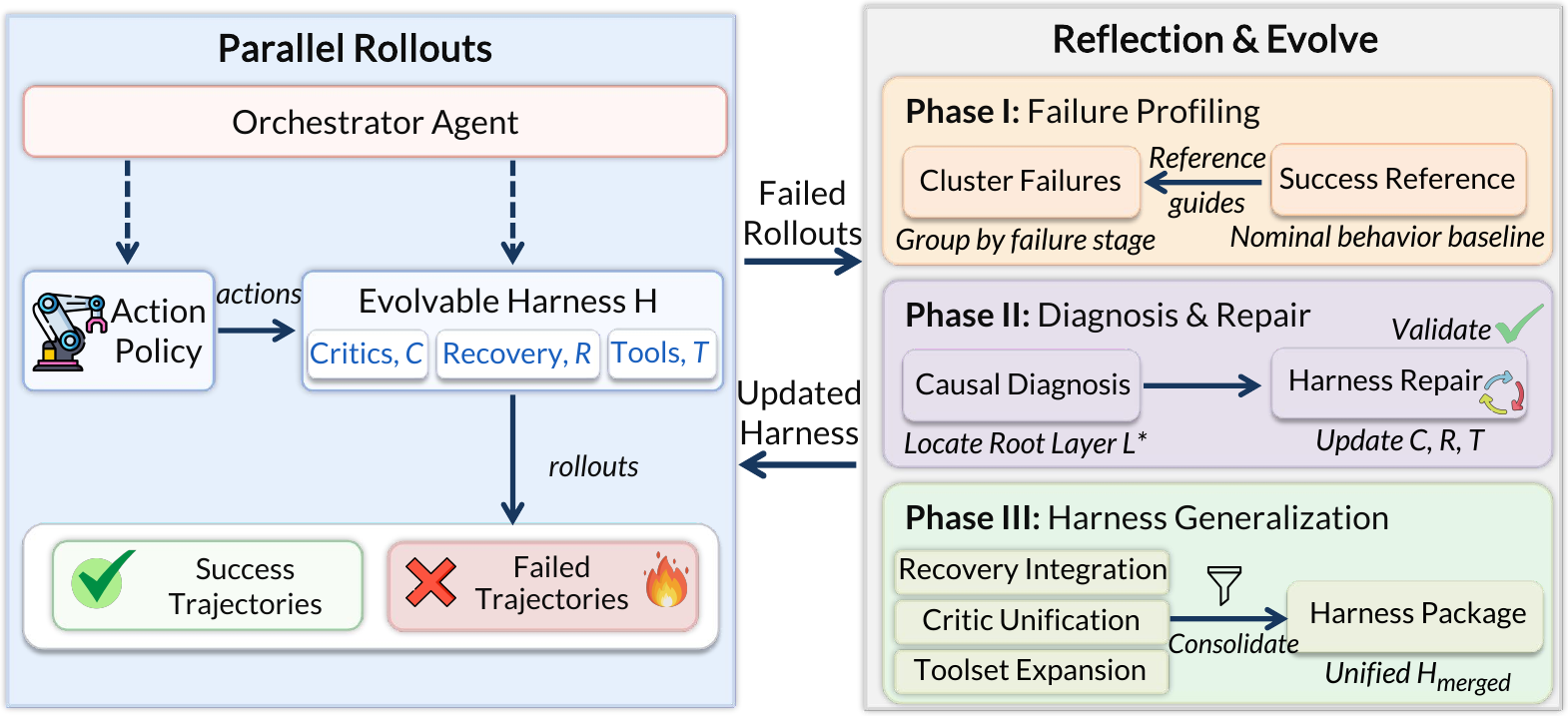}
 \caption{\textbf{Overview of the Zetta evolutionary framework.} The system operates in a continuous loop between online execution and offline evolution. \textbf{(Left) Parallel Rollouts:} Action policy executes tasks, monitored by an \textit{Evolvable Harness} ($\mathcal{H}$) composed of Critics ($C$), Recovery ($R$), and Tools ($T$), all under the high-level adjudication of an \textit{Orchestrator Agent}. Rollouts are categorized into success and failure trajectories. \textbf{(Right) Reflection \& Evolve:} Failed trajectories trigger a three-phase offline evolution cycle. \textit{Phase I} profiles failures by clustering them against successful reference baselines. \textit{Phase II} performs causal diagnosis to locate the root failure layer ($L^*$), followed by minimal harness repair to update $C$, $R$, and $T$. \textit{Phase III} generalizes seed-specific repairs into a unified, versioned \textit{Harness Package} ($\mathcal{H}_{merged}$), which is then fed back into the execution loop.}
    \label{fig:Zetta_overview}
\end{figure}

\subsection{System Challenges and Design Principles}

To address the limitations of current embodied AI systems based on frozen Vision-Language-Action (VLA) policies, we identify three core challenges that motivate the design of the Zetta framework.

\paragraph{Challenge 1: The Open-Loop Semantics-Physics Gap in Frozen Foundation Models.}
SOTA VLA/WAM models possess strong semantic understanding but operate as \textbf{open-loop} feedforward policies during inference \cite{black2025pi05,bjorck2025groot,li2026lingbotva,fastwam2026}. They lack closed-loop sensory-motor perception to correct physical execution errors (e.g., object slippage, minor collisions) in real-time. Consequently, minor disturbances often cascade into total task failure because the policy cannot monitor or adjust to its own physical state.

\noindent \textbf{Zetta Design Principle: High-Frequency State Governance.} We augment the frozen VLA/WAM with decoupled, \textbf{high-frequency runtime critics ($C$)}. Operating above the action policy's  inference rate, these critics continuously monitor the physical execution state and trigger interventions at the earliest signs of deviation from the nominal distribution, transforming the system into a closed-loop governance paradigm.

\paragraph{Challenge 2: The Generalization Pitfall of Over-Parameterized Repair.}
Determining the root cause of physical failures is complex. Ad-hoc debugging often leads to "overfitting" repairs—adjusting low-level control parameters to force success on a specific failure instance \cite{muratore2022robot,tan2018sim,rana2023bayesian}. While resolving the immediate fault, such modifications corrupt the action distribution essential for the VLA's semantic generalization, causing severe performance degradation on unseen held-out seeds, thus sacrificing global generalizability.

\noindent \textbf{Zetta Design Principle: Hierarchical Causal Diagnosis for Minimal Intervention.} We implement a strict \textbf{Top-Down Hierarchical Causal Diagnosis} logic. The Diagnosis Agent $\mathcal{A}_{diag}$ traverses layers in priority order: from $Evaluation \to Critic \to State \to Planning \to Recovery \to Parameter$.\cite{pettersson2005execution,donald1988geometric,gu2026safe} Adhering to the principle that ``if high-level logic resolves the failure, never modify low-level parameters,'' ensures patches are applied at the minimal effective layer, preserving the foundation model's integrity.

\paragraph{Challenge 3: The Scalability of Expert-in-the-Loop Debugging.}
Relying on human experts to manually analyze and patch every individual failure instance across long-tail task distributions is expensive and fundamentally unscalable \cite{hu2026rac,zhu2026beyond,li2025hamster,liu2024moka}. This approach cannot keep up with the infinite variations of initial conditions inherent in general-purpose manipulation.

\noindent \textbf{Zetta Design Principle: Automated Evolutionary Generalization.} We replace manual debugging with a \textbf{fully automated evolutionary learning loop (Loop 1-3)}. The Evolutionary Agent $\mathcal{A}_{evo}$ autonomously abstracts seed-specific failures into generalized harness based on \textit{task invariants}. Furthermore, an enforced \textbf{Held-out Generalization Protocol} validates consolidated harness on strictly isolated test sets, enabling automated scaling of robust governance strategies.

\subsection{Phase I: Empirical Failure Profiling and Baseline Establishment (Loop 1)}

At the onset of the evolutionary cycle, we conduct large-scale sampling of the Action policy $\pi$ across a predefined development seed set $\mathcal{D}_{dev}$. The objective is to establish a performance baseline through pure VLA closed-loop execution without governance intervention. The resulting raw corpus is defined as $\mathcal{B}_{raw} = \{ \tau^{(j)} \mid seed_j \in \mathcal{D}_{dev} \}$.

\paragraph{Deterministic Scheduling and Execution Protocol}
To ensure empirical rigor and reproducibility, Loop 1 enforces strict infrastructure management through two deterministic dimensions:
\begin{itemize}
    \item \textbf{Deterministic Routing}: All rollout tasks are dispatched via a centralized resource scheduler. The scheduler routes tasks based on real-time loads and memory thresholds of computation nodes (e.g., 4090 GPU pools). This mechanism ensures that all rollouts within a batch run under identical software containers, simulator versions, and hardware configurations, eliminating observational noise from environmental heterogeneity.
    \item \textbf{Validity Determination and Isolation}: The system explicitly distinguishes between \textit{infrastructure failure} and \textit{policy failure}. We define a validity function $\text{Valid}(seed_j) \in \{True, False\}$. A rollout is included in the \textbf{valid rollout set $\mathcal{V}$} only if it completes with a full trace of sensor data and video evidence:
    \begin{equation}
        \mathcal{V} = \{ seed_j \in \mathcal{D}_{dev} \mid \text{Valid}(seed_j) = True \}
    \end{equation}
    For infrastructure-invalid attempts (e.g., network fluctuations, simulator crashes), a mandatory rerun policy is enforced using the original logical seeds until a valid trajectory is produced, ensuring the statistical distribution of $\mathcal{V}$ does not drift due to non-policy factors.
\end{itemize}

\paragraph{Multi-dimensional Evidence Acquisition}
For each valid rollout $\tau^{(j)} \in \mathcal{V}$, the system employs an \textbf{append-only} mode to preserve a comprehensive observational stream. A complete trajectory instance $\tau^{(j)}$ is represented as a multimodal time series:
\begin{equation}
    \tau^{(j)} = \{ (s_t, a_t, \mu_t, \phi_t) \}_{t=0}^T
\end{equation}
where $\mu_t$ denotes the completion status of task-specific semantic milestones, and $\phi_t$ captures physio-auxiliary signals such as collision intensities and contact force vectors.

\paragraph{Categorization and Indexing}
The resulting corpus is processed into two core repositories for the Evolutionary Agents $\mathcal{A}_{evo}$:
\begin{enumerate}
    \item \textbf{Successful Reference Index ($I_{succ}$)}: Successful trajectories $\mathcal{V}_{succ} \subset \mathcal{V}$ are aggregated by milestones:
    \begin{equation}
        I_{succ}(\mu) = \{ s_t \mid \tau \in \mathcal{V}_{succ}, \mu_t = \mu \}
    \end{equation}
    This index reveals the ``nominal distribution'' of task success, serving as a benchmark for identifying deviations.
    \item \textbf{Failed-Seed Manifest ($\mathcal{M}_{fail}$) and First Missing Milestone ($m^*$)}: All unsuccessful samples and their evidence chains are compiled into a manifest. To localize the task stage where failure occurred, we introduce the \textbf{First Missing Milestone ($m^*$)}.
    
    Given an ordered sequence of semantic milestones $M = \langle m_1, m_2, \dots, m_{goal} \rangle$, $m^*$ is defined as the first element in the sequence that was not observed in the trajectory history:
    \begin{equation}
        m^* = \min \{ m_k \in M \mid m_k \notin \{ \mu_t \}_{t=0}^T \}
    \end{equation}
    The introduction of $m^*$ serves two purposes: providing a coarse-grained clustering criterion for failures and narrowing the search space for causal diagnosis by focusing $\mathcal{A}_{diag}$ on the evidence during the transition from $m^*-1$ to $m^*$.
\end{enumerate}

\noindent \textbf{The Evolution Cycle: From Data to Harness.} Based on the formalization of the Evolutionary Agent $\mathcal{A}_{evo}$ in Section \ref{sec:problem}, the operational logic is realized through a tripartite pipeline: \textbf{Diagnosis} ($\mathcal{A}_{diag}$), \textbf{Repair} ($\mathcal{A}_{repr}$), and \textbf{Generalization} ($\mathcal{A}_{gen}$). This structure ensures the systematic extraction of causal knowledge from the raw failure corpus $\mathcal{M}_{fail}$. As the pipeline's starting point, the primary mission of $\mathcal{A}_{diag}$ is to identify the root causal mechanisms of failures.

\subsection{Phase II Stage 1: Failure Clustering and Causal Diagnosis}

\subsubsection{Mechanism-Level Failure Clustering and Medoid Seed Selection}
The diagnosis begins with a structural analysis of the Failed-Seed Manifest $\mathcal{M}_{fail}$ produced in Loop 1. Instead of superficial categorization based on final outcomes, $\mathcal{A}_{diag}$ clusters seeds based on the \textbf{Earliest Observable Divergence (EOD)}. We define $t_{EOD}$ as the first time step in a trajectory $\tau \in \mathcal{M}_{fail}$ where the state distribution deviates from the ``healthy'' distribution described by $I_{succ}$:
\begin{equation}
    t_{EOD} = \min \{ t \mid \text{dist}(s_t, s_t^{ref}) > \epsilon, s_t^{ref} \in I_{succ}(\mu_t) \}
\end{equation}
where $\text{dist}(\cdot)$ is a state-space distance metric and $\epsilon$ is a predefined threshold. 

Utilizing $t_{EOD}$ and its context—including the first missing milestone $m^*$, robot-object relative poses, and active tool IDs—$\mathcal{A}_{diag}$ partitions $\mathcal{M}_{fail}$ into mutually exclusive clusters $\mathcal{K} = \{K_1, K_2, \dots, K_n\}$, where each $K_i \subseteq \mathcal{M}_{fail}$ contains seeds sharing similar failure signatures. To reduce computational overhead and extract mechanism invariants, a \textbf{medoid seed} $seed_{med} \in K_i$ is selected for each cluster. Defined as the sample closest to the cluster center in the feature space, $seed_{med}$ serves as the primary subject for in-depth causal diagnosis.

\paragraph{Single-View Grounded Observation Protocol}
To eliminate spatial reasoning inconsistencies and multiview hallucinations \cite{wu2026grounded,dongfang2026multimodal,huang2022inner} in multimodal models, $\mathcal{A}_{diag}$ adheres to a strict \textbf{single-view grounded protocol}:
\begin{itemize}
    \item \textbf{Primary View Selection}: The agent must first designate a primary view from available video streams, defined as the angle that most clearly exhibits the end-effector, the target object, and the critical contact interface.
    \item \textbf{Evidence Consistency}: Once the primary view is established, all visual reasoning throughout the diagnosis must be anchored to it. If the primary view provides insufficient evidence, the system is mandated to revert to internal simulator states, sensor trajectories, and physical signals $\phi_t$ for cross-modal verification, rather than switching viewpoints, ensuring spatial-logical consistency.
\end{itemize}

\subsubsection{Top-Down Hierarchical Causal Diagnosis}
For the selected $seed_{med}$, $\mathcal{A}_{diag}$ executes a systematic top-down inspection logic across the diagnostic layer space $\mathcal{L} = \{L_{eval}, L_{crit}, L_{state}, L_{plan}, L_{recv}, L_{param}\}$. The task is to localize the root cause by finding the highest-priority layer $L^*$ such that:
\begin{equation}
    L^* = \text{arg max}_{L \in \mathcal{L}} \{ \text{IsRootCause}(L) \mid \text{Evidence from } \tau, \text{Primary View} \}
\end{equation}
The inspection order follows the hierarchical priority:
\begin{enumerate}
    \item \textbf{Evaluation Layer ($L_{eval}$)}: Check for errors in success criteria or milestone progress logic.
    \item \textbf{Critic Layer ($L_{crit}$)}: Determine if runtime critics $C$ exhibit false negatives or false positives.
    \item \textbf{State Representation Layer ($L_{state}$)}: Verify if object poses or contact information deviated from physical ground truth.
    \item \textbf{Planning/Control Layer ($L_{plan}$)}: Analyze if the VLA policy or tools failed to handle physical constraints despite correct states.
    \item \textbf{Recovery Layer ($L_{recv}$)}: If the failure occurred during recovery, check for flaws in the playbook logic.
    \item \textbf{Parameter Layer ($L_{param}$)}: Inspect specific control gains or action thresholds.
\end{enumerate}

\textbf{Cluster Consistency and Repair Specification}
Upon determining $L^*$ for $seed_{med}$, $\mathcal{A}_{diag}$ verifies the mechanism's consistency across other members of $K_i$. Finally, the agent outputs a \textbf{Repair Candidate Specification} for each validated mechanism, detailing the localized layer, the causal evidence chain, and functional requirements for updating the harness components ($C, R, \mathcal{T}$).

\subsection{Phase II Stage 2: Critic-Guided Harness Repair and Validation}
\label{sec:phase_ii_stage_2}

This stage is led by the Repair Agent $\mathcal{A}_{repr}$, whose objective is to implement a minimal harness patch $\mathcal{H}_{patch} = \{C^*, R^*, \mathcal{T}^*\}$ that resolves the specific root layer $L^*$ identified during diagnosis, while strictly preserving the integrity of the base Action Policy $\pi$.

\subsubsection{Harness Component Instantiation and Re-entry Logic}
$\mathcal{A}_{repr}$ begins by instantiating the requisite governance components dictated by the diagnostic specification. This entails targeted updates across the harness:
\begin{itemize}
    \item \textbf{Critic Enhancement ($C^*$)}: Development or refinement of high-frequency monitoring functions designed to detect the specific precursor conditions of the failure, generating structured proposals $P_t$ upon deviation.
    \item \textbf{Recovery Drafting ($R^*$) and Tool Adaptation ($\mathcal{T}^*$)}: Definition of actionable recovery playbooks and the adaptation or synthesis of executable tools to address the causal failure mechanism.
\end{itemize}

Crucially, to \textbf{ensure safe and seamless integration}, $\mathcal{A}_{repr}$ embeds a strict \textbf{VLA Re-entry Contract} within the recovery logic of $R^*$ and $\mathcal{T}^*$. This contract defines the logical predicate $\Psi(s_t)$ that dictates when control is relinquished back to the base policy $\pi_{VLA}$. Control handover is permitted only if the original failure evidence $e_t$ is explicitly cleared and the physical state has reached equilibrium, as defined by stable contact forces. We define this predicate as:
\begin{equation}
    \Psi(s_t) = \mathbb{1}(\text{FailureCleared}) \land \mathbb{1}(\text{Stability}(s_t) > \gamma)
\end{equation}
where the \textbf{Failure Cleared} term is a boolean function verifying that the specific conditions constituting evidence $e_t$ (e.g., collision risk or pose deviation) have been fully resolved by the recovery actions. Simultaneously, the \textbf{Stable Contacts} term evaluates physio-auxiliary signals $\phi_t$ to ensure the connection between the robot and the object has reached a physical equilibrium. Specifically, $\text{Stability}(s_t)$ measures the magnitude of contact torque oscillations and grasp forces, while $\gamma$ is a stability threshold ensuring safe takeover by the VLA, preventing secondary failures due to transient dynamic effects. This contract acts as a crucial safeguard, preventing instability from triggering secondary failures upon policy resumption.

\subsubsection{Closed-Loop Validation Protocol and Success Criteria}
Following implementation, the patch $\mathcal{H}_{patch}$ undergoes a rigorous two-step closed-loop validation on the original medoid seed. First, a \textit{diagnostic replay} confirms that the enhanced critic $C^*$ now correctly identifies the divergence point $t_{EOD}$. Second, a \textit{fresh closed-loop rollout} is executed from the initial state. A repair patch is only considered validated and successfully \textit{passed} if the task reaches the goal milestone and all interventions are properly adjudicated by $\mathcal{A}_{orch}$ in compliance with the re-entry contract:
\begin{equation}
    \text{Success}(\mathcal{H}_{patch}) = \mathbb{1}(\mu_{T, new} = m_{goal} \land \forall t \in \text{Intv}, \text{Adjudicated by } \mathcal{A}_{orch})
\end{equation}
where $\mu_{T, new}$ is the final milestone of the new trajectory and $\text{Intv}$ represents the intervention interval. Validated patches, along with their evidence, are then submitted to $\mathcal{A}_{gen}$ for multi-seed consolidation.

\subsection{Phase III: Harness Consolidation, Packaging and Generalization}
\label{sec:phase_iii}

In this final stage of the evolutionary cycle, the Generalization Agent $\mathcal{A}_{gen}$ transforms validated, seed-specific patches into a robust, unified governance harness. The objective is to abstract local repairs into mechanism-level invariants capable of resolving the entire failure cluster $K_i$, resulting in a versioned \textbf{merged harness $\mathcal{H}_{merged}$}.

\subsubsection{Mechanism Consolidation and Harness Packaging}
$\mathcal{A}_{gen}$ performs a cross-seed analysis to consolidate heterogeneous patches $\mathcal{H}_{patch, j}$ (for $seed_j \in K_i$) into a single, coherent configuration $\mathcal{H}_{merged} = \{C_{merged}, R_{merged}, \mathcal{T}_{merged}\}$. This process elevates governance logic from specific instances to general mechanisms and encapsulates it into a portable, standardized file structure.

The consolidation operator abstracts local experiences into task invariants:
\begin{itemize}
  \item \textbf{Critic Unification}: Individual monitoring functions are abstracted into a unified Critic $C_{merged}$. By applying logical OR operations on triggering conditions and dynamically adjusting noise thresholds, $C_{merged}$ ensures reliable failure detection across the diverse initial conditions present within the cluster.
    \item \textbf{Recovery Integration}: Seed-specific Recovery Playbooks are abstracted into $R_{merged}$, which generalizes action sequences to handle morphological variations of the same failure mode and standardizes the VLA Re-entry Contract $\Psi(s_t)$.
    \item \textbf{Toolset Expansion}: The Heterogeneous Toolset $\mathcal{T}_{merged}$ is expanded to include not only tuned existing operators but also \textbf{newly synthesized tools}. These executable scripts encapsulate specialized physical skills (e.g., high-precision impedance control) generated during the repair phase to address constraints beyond the base VLA's capabilities.
\end{itemize}
Mathematically, this consolidation is defined as:
\begin{equation}
    \mathcal{H}_{merged} = \text{Consolidate} \left( \{ \mathcal{H}_{patch, j} \mid seed_j \in K_i \} \right)
\end{equation}

To ensure the portability and version control of these evolved capabilities, $\mathcal{A}_{gen}$ externalizes the governance logic into a self-contained package, mapping components to distinct directories:
\begin{itemize}
    \item \textbf{$SKILL.md$}: Declares the high-level governance logic, including defined milestones $M$, Orchestrator adjudication rules, and the generalized re-entry predicate $\Psi(s_t)$.
    \item \textbf{$tools/$}: Contains the executable implementations and scripts for the evolved critics $C_{merged}$ and the toolset $\mathcal{T}_{merged}$.
    \item \textbf{$plans/$}: Stores the generalized recovery playbooks $R_{merged}$, mapping specific evidence to structured intervention strategies.
    \item \textbf{$params/$}: Holds the configuration files defining generalized numeric boundaries and thresholds for runtime execution.
\end{itemize}

\subsubsection{Rigorous Generalization Validation and Transition Protocol}
The final validated harness $\mathcal{H}_{merged}$ must satisfy stringent performance criteria via a dual-evaluation protocol to confirm genuine generalization.

\paragraph{Historical Regression and Held-Out Evaluation}
First, the harness undergoes **Historical Regression** testing, requiring it to successfully resolve all failed rollouts within the originating cluster $K_i$ with a 100\% success rate:
\begin{equation}
    \forall seed_j \in K_i, \text{Success}(seed_j \mid \mathcal{H}_{merged}) = 1
\end{equation}
Second, to confirm broader applicability, the harness is subjected to **Held-out Evaluation** on a strictly isolated dataset $\mathcal{D}_{held-out}$. The effectiveness of the evolution is quantified by the success rate increment $\Delta SR$ over the baseline VLA policy:
\begin{equation}
    \Delta SR = SR(\mathcal{D}_{held-out} \mid \mathcal{H}_{merged}) - SR(\mathcal{D}_{held-out} \mid \pi_{VLA})
\end{equation}

\paragraph{Dynamic Transition Protocol}
An evolutionary iteration is considered complete only when $\mathcal{H}_{merged}$ demonstrates robust performance on previously unseen data. If the evaluation on $\mathcal{D}_{held-out}$ reveals novel failure mechanisms that trigger modifications to $\mathcal{H}_{merged}$, the system enforces a strict data segregation policy. The original held-out seed is reclassified as a development seed ($\mathcal{D}_{dev} \leftarrow \mathcal{D}_{dev} \cup \{seed_{failed}\}$), and a fresh, previously unseen set must be selected for final validation before the iteration can be closed.

\section{Z-Infra: Embodied Agent Rollout Infrastructure}

\subsection{Overview}

The rollout infrastructure serves as the execution backbone for the self-evolving embodied agent described above.
Its primary responsibility is to efficiently execute large-scale parallel rollouts (complete episodes of agent-environment interaction) across heterogeneous compute resources while exposing a simple, unified interface to upper-layer agent logic.

A single rollout proceeds as follows.
The agent first requests a \emph{session}, which provisions an isolated environment instance on an appropriate simulation environment, and then issues a \texttt{reset} call to initialize an episode.
The main execution loop iterates: the agent observes the environment state and invokes \texttt{policy\_step} to query a policy model for actions, or calls perception models and primitive operations as needed.
This loop continues until termination (success, failure, or timeout), at which point the session is released.

The self-evolving agent runs \emph{many such rollouts concurrently}, exploring different tasks, testing learned skills, and collecting experience for reflection, making efficient multiplexing of shared compute resources essential.
Although this work primarily considers the CPU-centric MuJoCo~\cite{todorov2012mujoco}/robosuite~\cite{zhu2020robosuite} stack used by LIBERO~\cite{liu2023libero} and RoboCasa~\cite{nasiriany2024robocasa}, embodied simulation also includes GPU-parallel MuJoCo backends such as MJX~\cite{mujocoteam2026mjx} and MJLab~\cite{zakka2026mjlab}, as well as PhysX-based stacks such as SAPIEN/ManiSkill~\cite{xiang2020sapien,gu2023maniskill2} and Isaac Sim/Isaac Lab~\cite{gao2026nvidia,mittal2025isaac}.
The rollout interface is designed to accommodate these backend differences while presenting the same session-level interaction model to the agent.

\subsection{Challenges and Key Ideas}

The agentic rollout workload described above poses two fundamental infrastructure challenges:

\paragraph{Resource Heterogeneity.}
A single rollout simultaneously demands diverse compute resources.
For the CPU-centric simulations, MuJoCo/robosuite environments maintain per-session simulation state and consume host CPU and memory during environment stepping, while their conventional rendering pipelines commonly use GPUs.
GPU-parallel simulation backends instead advance batches of environments on accelerators, with different requirements for state layout, reset semantics, rendering, and device memory.
Policy models require GPU accelerators for autoregressive or diffusion-based action generation and GPU memory to store computing feature~\cite{li2026oxygen,xu2026embodied,ma2025runningvlasrealtimespeed,tang2025vlash}.
Lightweight perception models~\cite{carion2025sam3segmentconcepts,kirillov2023segment,Jocher_Ultralytics_YOLO26_Unified_2026,redmon2016you,ding2025streammind,zheng2026emgardeproposematchframeworkproactive,fang2023robust,sundermeyer2021contact} can share GPU resources but have distinct latency requirements.
Primitive operations (coordinate transforms, collision checks) are pure CPU computations with microsecond-scale latency expectations.
These workloads lead to different worker and scheduling requirements, so no single resource pool or scheduling policy fits all these workload types.

\paragraph{Execution Dynamism.}
Unlike training pipelines with predictable data-parallel patterns, agentic rollouts exhibit inherently unpredictable execution profiles.
The agent dynamically selects which tools to invoke based on runtime observations, one timestep may require only a policy call, while the next may chain perception, planning, and multiple primitive operations.
Sessions are created, paused, and destroyed at irregular intervals as the agent explores tasks, reflects on failures, and retries with modified strategies.
This results in bursty GPU inference demands with high variance in batch sizes and inter-arrival times, making static resource allocation and request scheduling inefficient.

\paragraph{Key Idea: Decoupled Abstraction Layers.}
We address these challenges by introducing an abstraction layer that fully decouples agent logic from hardware resource management.
The agent interacts exclusively with a virtual rollout interface---specifying \emph{what} to execute (which model, which environment, which operation) without concern for \emph{where} or \emph{how} execution occurs.
The infrastructure independently handles worker allocation, request routing, batch formation, and hardware mapping, enabling transparent scaling across multi-node, multi-GPU clusters without requiring changes to agent code.
This separation of concerns allows each layer to be optimized independently: agent developers focus on task logic and skill composition, while infrastructure engineers optimize scheduling, batching, and hardware utilization.

\subsection{System Architecture}

\begin{figure}[t]
  \centering
  \includegraphics[width=0.72\linewidth]{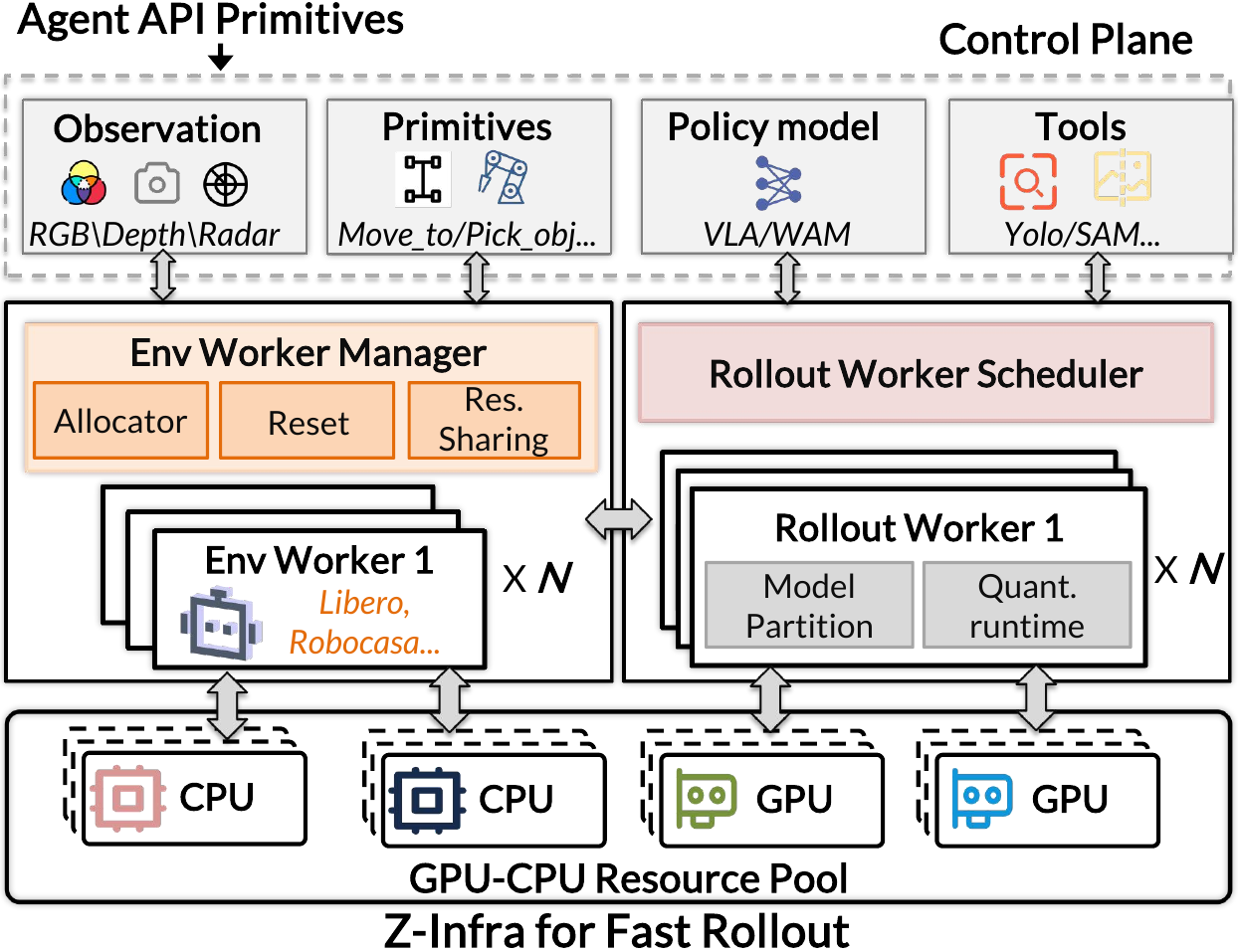}
  \caption{Three-layer architecture of the rollout infrastructure. The Control Plane routes agent requests to specialized Env Workers and Rollout Workers, which manage environment simulation and model inference respectively.}
  \label{fig:infra-architecture}
\end{figure}

The infrastructure is organized into three layers (Figure~\ref{fig:infra-architecture}), each addressing a distinct concern in the execution pipeline.
The \textbf{Control Plane} serves as the single entry point for all agent interactions, exposing a unified API that abstracts over backend heterogeneity and handles request routing, resource management, and fault tolerance.
The \textbf{Environment Worker Layer} manages the lifecycle and execution of simulation environments across diverse families, handling session provisioning, environment stepping, and observation capture.
The \textbf{Rollout Worker Layer} provides GPU-resident model serving for VLA/WAM policies and perception models, implementing batched inference with dynamic scheduling to maximize throughput under variable request patterns.
These layers communicate via bounded asynchronous channels that enforce backpressure and enable transparent scaling across multi-node GPU clusters.

\subsubsection{Control Plane}

The Control Plane serves as the infrastructure's coordination hub, decoupling agent logic from distributed resource management.
It exposes a unified API that routes each request to the appropriate worker pool based on request type (environment operation vs. model inference) and resource requirements (environment family, model type).
The Gateway maintains a global session registry that tracks the binding between sessions and Env Worker ranks, enabling efficient request dispatch without requiring agents to manage worker topology.
Worker health is monitored via periodic heartbeats: when an Env Worker fails, the Gateway detects the timeout, marks affected sessions as \texttt{LOST}, and returns failure notifications to agents, which can then recreate sessions on healthy workers.


\subsubsection{Environment Worker}

The Environment Worker layer addresses the challenge of managing heterogeneous simulation environments at scale while maintaining session isolation and maximizing resource utilization.
Each Env Worker is a distributed actor that hosts multiple environment slots, where each slot binds to an active session and maintains its own execution state.
The layer must support diverse environment families with different APIs (reset signatures, observation structures, action formats) and execution models (CPU subprocesses vs. GPU-batched simulation), while exposing a uniform interface to the Control Plane.

\paragraph{Session-based Lifecycle Management.}
We introduce a session abstraction to support long-lived agent-environment interactions.
Each session encapsulates an independent execution context, binding to a specific environment slot on an Env Worker and progressing through a well-defined lifecycle: \texttt{CREATE} $\rightarrow$ \texttt{RUNNING} $\rightarrow$ \texttt{TERMINATED}.
Upon creation, the Gateway selects an appropriate Env Worker based on resource availability and environment family compatibility, allocates a slot, and returns a session handle to the agent.

The session registry maintained by the Gateway enables efficient resource accounting and admission control.
At any point, the system knows exactly how many sessions of each environment family are active, their distribution across workers, and remaining lease durations.
When workers are saturated, new session requests are rejected with backpressure signals, preventing cascading overload.

To onboard new environment families, developers implement four primitives through a declarative registration framework:
\textbf{(i)} \texttt{init}---construct the environment from a configuration dictionary;
\textbf{(ii)} \texttt{reset}---reset to an initial state given task-specific parameters;
\textbf{(iii)} \texttt{step}---execute an action and return observation, reward, and termination flags;
\textbf{(iv)} \texttt{obs\_schema}---declare the structure and encoding of observations.
A normalization layer translates family-specific observations into a canonical format (dictionaries of named tensors with declared shapes and dtypes), ensuring that upper-layer code remains family-agnostic.

\paragraph{Resource-Sharing Group.}
For parallel benchmark rollouts using MuJoCo-based environments (e.g., LIBERO~\cite{liu2023libero}, RoboCasa~\cite{nasiriany2024robocasa}), we implement a resource-sharing optimization that amortizes model compilation and rendering context setup across multiple sessions.
The resource-sharing group is the execution unit for sessions with identical environment structure and configuration.
The group compiles the environment model once into a \texttt{ModelTemplate}, which retains the reusable read-only \texttt{mjModel} representation together with the initial simulation state.
Each slot is initialized through a fork primitive from this template, allowing the group to reuse immutable model resources while each slot maintains its own \texttt{mjData} and episode state.
This separates shared model structure from mutable simulation state: resetting or stepping one slot does not modify the state of any other slot in the group.

The group also owns the execution resources required by the slots.
Each slot is assigned to a pinned worker thread with a thread-local render context, while the contexts belong to a shared render context group, allowing compatible GPU rendering resources to be reused across slots without cross-thread context conflicts.
To make this resource sharing effective under multithreaded execution, the environment step routine is reimplemented with a high-performance C++ controller that combines control execution, action interpolation, physics substeps, and rendering, while minimizing Python-side overhead.
With the GIL released during the critical path, multiple slots advance their physics and rendering workloads in parallel within the same worker process.

\subsubsection{Rollout Worker}

The Rollout Worker layer addresses the challenge of serving diverse neural models under dynamic, bursty request patterns while maximizing GPU utilization.
Each Rollout Worker is a GPU-resident actor that executes batched inference and returns results asynchronously.
Workers are organized into specialized pools: policy models occupy dedicated high-memory GPUs, while lightweight perception models are co-located on shared GPUs.
The layer must balance throughput (larger batches amortize GPU kernel overhead) against latency (requests should not wait indefinitely), while handling compatibility constraints (only requests with matching model type, input modalities, and tensor shapes can be batched together).

\paragraph{Scheduler.}
The Rollout Worker Scheduler manages the full lifecycle of inference requests from arrival to result delivery across a pool of GPU-resident workers.
Upon receiving an inference request, the scheduler first classifies it into a compatibility group based on model identity, input modalities, and tensor shapes, then enqueues it into the corresponding per-group request queue.
The scheduler monitors worker load and dispatches batched requests to available workers in a first-come-first-served (FCFS) manner to ensure high overall resource utilization while maintaining low per-request latency.
When a Rollout Worker completes a batch, it tags each result with its routing token and publishes it to a shared result channel; the originating Env Worker retrieves its results by token match, enabling fully asynchronous request-response patterns without blocking.

\paragraph{Model Partitioning.}
Modern VLA/WAM architectures (e.g., $\pi_{0.5}$~\cite{black2025pi05}) consist of two functionally distinct stages: a Vision-Language Model (VLM) that encodes observations and instructions into latent representations, and an Action Expert (AE) that decodes these representations into motor commands via diffusion or autoregressive generation.
These stages have markedly different computational characteristics.
We exploit these different characteristics by deploying the VLM and AE as separate processes with independent scheduling policies.
VLM's intermediate activations are transferred between VLM and AE processes via CUDA IPC to avoid expensive transmission overhead.
For the $\pi_{0.5}$ model implemented in PyTorch, this partitioning strategy reduces average inference latency by 53\% and improves goodput (under 200ms SLO) by 2.4$\times$ compared to monolithic deployment.

\paragraph{Quantization Runtime.}
To improve policy inference throughput and GPU memory efficiency, we introduce quantization (e.g., W8A8, W4A16, W4A8) as an \emph{optional} plug-in to the Rollout Worker.
Unlike model-level quantization in common LLM serving frameworks~\cite{kwon2023efficient,zheng2024sglang}, we quantize different modules inside a policy model separately, co-designed with model partitioning, as well as to balance policy success rates and inference efficiency. For $\pi_{0.5}$ on RTX 4090, we currently apply W8A8 quantization for the  compute-intensive prefix MLP modules, while keeping other modules at BF16 to avoid additional overhead, achieving $1.18\times$--$1.32\times$ inference speedup without degrading success rates on LIBERO.

\subsubsection{Implementation.}
The infrastructure is built on Ray~\cite{222605} for distributed execution.
Each Env Worker and Rollout Worker is a Ray actor wrapping the respective runtime logic, with resource annotations declared via placement strategies.
We use Ray's placement strategy to bind each worker to a dedicated GPU/CPU.
For example, declaring \texttt{placement: "0-7"} provisions 8 ranks, each mapped to one GPU (e.g., 8 A100s on a single node, or distributed across nodes in multi-node setups).
Each rollout rank loads an independent copy of the model, enabling data-parallel inference across the rank pool.

Communication across the three layers is structured around five bounded Ray Channels that partition control flow and data flow.
Command and control channels route lifecycle operations and priority signals from Gateway to Env Workers, while result channels return step outcomes in the reverse direction.
Inference requests flow from Env Workers to a shared channel consumed competitively by all Rollout Workers, enabling load-aware distribution; responses are routed back via embedded tokens.

\subsection{Programming Support}

The infrastructure exposes a layered programming model designed to minimize cognitive overhead for agent developers while retaining full flexibility for advanced use cases.
The model is organized around three core abstractions: \textbf{sessions} represent long-lived environment instances with managed lifecycles, \textbf{episodes} run within sessions from reset to termination, and \textbf{steps} execute individual actions within episodes.
Agents interact through a clean API that handles batching, asynchronous inference, and resource management transparently---developers specify \emph{what} they want (e.g., ``execute this policy in these environments''), and the infrastructure determines \emph{how} to schedule workers, batch requests, and route results efficiently.

\paragraph{API Overview.}
Table~\ref{tab:api-primitives} summarizes the core API exposed to upper-layer agents, organized into five categories: \textit{Session Management} handles environment lifecycle (create, renew, close); \textit{Observation \& State} provides environment introspection; \textit{Low-Level Control} executes motor commands (Cartesian servoing, gripper control); \textit{Policy Execution} integrates model serving (\texttt{policy\_step} for atomic observe-infer-step operations, \texttt{run\_episode} for autonomous execution); and \textit{Perception \& Planning} provides high-level services (segmentation, grasp generation, motion planning).
All APIs support transparent batching across multiple sessions.

\begin{table}[t]
\centering
\small
\begin{tabular}{@{}p{0.38\linewidth}p{0.58\linewidth}@{}}
\toprule
\textbf{API Primitive} & \textbf{Functionality} \\
\midrule
\multicolumn{2}{@{}l}{\textit{Session Management --- create once, run many episodes, close when done}} \\
\texttt{create\_sessions(requests)} & Batch create sessions with env\_family, env\_config, lease\_seconds. Returns SessionHandle with session\_id. \\
\texttt{renew\_sessions(session\_ids)} & Extend lease duration for active sessions. \\
\texttt{close\_sessions(session\_ids)} & Close sessions and release resources (idempotent). \\
\midrule
\multicolumn{2}{@{}l}{\textit{Environment Control --- direct interaction for custom control logic}} \\
\texttt{reset(session\_ids, reset\_spec)} & Start new episode with task\_id, seed, instruction. Returns episode\_id and initial observation. \\
\texttt{observe(session\_ids)} & Read current observation without side effects. \\
\texttt{action\_step(session\_ids, actions)} & Execute actions, return observation, reward, terminated, truncated flags. \\
\midrule
\multicolumn{2}{@{}l}{\textit{Policy Inference --- integrated model serving and stepping}} \\
\texttt{policy\_step(session\_ids, policy\_req)} & Atomic observe $\rightarrow$ inference $\rightarrow$ step. Returns step results with executed actions. \\
\texttt{policy\_infer(session\_ids, policy\_req)} & Inference only (no stepping). Returns actions for agent post-processing. \\
\texttt{run\_episode(session\_ids, episode\_req)} & Execute complete episodes in workers. Returns summary (steps, reward, stop\_reason). \\
\bottomrule
\end{tabular}
\caption{Core API primitives exposed to agents. All primitives support batching across sessions.}
\label{tab:api-primitives}
\end{table}

\begin{lstlisting}[language=Python, style=rolloutflow, caption={Simplified pseudocode for a \texttt{policy\_step} call spanning the Agent, Gateway, EnvWorker, and RolloutWorker layers.}, label=lst:execution-flow, captionpos=b, float, floatplacement=t]
# === Agent Side ===
sessions = await gateway.create_sessions(env_family="maniskill", n=16)
obs = await sessions.reset(task_id="pick_cube")
for _ in range(max_steps):
    result = await sessions.policy_step(instruction="pick_up_the_red_cube")
await sessions.close()

# === Gateway: policy_step dispatch ===
async def policy_step(session_id, instruction):
    worker_rank = registry.get_worker(session_id)
    return await send_command(worker_rank, "POLICY_STEP",
                               session_id=session_id, instruction=instruction)

# === EnvWorker: handle policy_step ===
async def handle_policy_step(session_id, instruction):
    token = make_routing_token(self.rank, session_id)
    await infer_req_ch.put(observation=session.obs, instruction=instruction,
                           routing_token=token)       # non-blocking, yield
    action = await infer_resp_ch.get(key=token)       # wake on response
    outcome = env.step(action)
    return StepResult(outcome)

# === RolloutWorker: batched inference ===
async def serve():
    while True:
        batch = await scheduler.next_batch()          # work-stealing
        actions = model.infer([r.obs for r in batch],
                              [r.inst for r in batch])
        for req, action in zip(batch, actions):
            await infer_resp_ch.put(key=req.routing_token, action=action)
\end{lstlisting}

\paragraph{Execution Flow.}
Listing~\ref{lst:execution-flow} illustrates a \texttt{policy\_step} call across the three infrastructure layers.
From the agent's perspective, the interaction is simple: create a session, reset the episode, and loop over \texttt{policy\_step}.
Internally, the Gateway routes each step to the session's Env Worker, which packages an inference request with a routing token and submits it to the shared channel before yielding to serve other sessions.
A Rollout Worker pulls the request, executes batched inference, and routes the result back via the token.
The Env Worker then steps the environment and returns the outcome, completing the pipeline while other sessions on the same worker continue making progress.

%% file: tex/experiment.tex
\newpage

\section{Experiments}

\subsection{Experimental Setup}
\label{sec:4_setup}

We evaluate the framework on LIBERO-Pro \cite{libero_pro} and RoboCasa \cite{nasiriany2024robocasa} benchmarks using a high-performance cluster equipped with \textbf{8 NVIDIA GeForce RTX 4090 GPUs}, enabling parallel rollouts and accelerated offline evolution.

For these experiments, we utilize specific pre-trained models as the underlying frozen base policies:
\begin{itemize}
    \item \textbf{LIBERO-Pro Benchmark}: Employs the $\pi_{0.5}$ \cite{black2025pi05} model as the base policy.
    \item \textbf{RoboCasa Benchmark}: Employs the GR00T N1.5 \cite{bjorck2025groot} model as the base policy.
\end{itemize}
Throughout all experiments, deployment-time evolution does not involve fine-tuning the weights of these base VLA models.

For both RoboCasa and LIBERO-Pro, we adhere to a \textbf{strict generalization evaluation protocol}: upon completion of the evolution, the final harness is evaluated on a \textbf{separate, strictly isolated set of test seeds} that were never seen by the Evolutionary Agent during the entire evolution and development process.

Specific implementation details for each benchmark are as follows:

\paragraph{RoboCasa: Random-Distribution Generalization.}
This evaluates the framework's generalization capabilities against challenging long-tail distributions, utilizing randomly sampled seeds for both development and testing.
\begin{enumerate}
    \item \textbf{Setup}: For each task, we randomly sample 50 environment seeds for development and evolution.
    \item \textbf{Evolutionary and Repair Cycle}: In each round, we collect failure trajectories from these 50 seeds and cluster them based on failure signatures. For each failure cluster, we select the representative medoid seed for diagnosis and repair development.
    \item \textbf{Final Evaluation (Held-Out)}: Upon completion, the final evaluation is conducted on a separate, strictly isolated set of 50 RoboCasa environment seeds (disjoint from the development set) to report the definitive success rate.
\end{enumerate}
\paragraph{LIBERO-Pro: Development-Test Generalization.}
This tests the framework's ability to generalize from a development set to entirely unseen task environments.
\begin{enumerate}
    \item \textbf{Setup}: For each task, we randomly sample 50 development seeds (Parent seeds, excluding seeds 1-20) for evolution.
    \item \textbf{Evolutionary Cycle and Repair}: Based on clustering of trajectories from the 50 development seeds, we target the erroneous seeds from the maximum failure cluster for repair development. Iteration continues until the success rate on these specific development seeds reaches $\ge 50\%$.
    \item \textbf{Final Evaluation (Held-Out)}: Upon completion, the final evaluation is conducted exclusively on the strictly isolated seeds 1 through 20 to report the SR.
\end{enumerate}

\subsection{Physical Intelligence ``Aha'' Moments}

To validate that Z-Harness enables discontinuous capability gains through physical intelligence rather than additional policy training, we analyze the fine-grained evolutionary trajectory within individual outer-loop cycles. We present \textbf{``Aha'' moments}—instances where the system transitions from stagnant performance to high success rates by identifying and resolving the true physical bottleneck of a task.

We define an ``Aha'' moment empirically by tracking success rates across \textbf{selected cumulative internal versions} within the evolution of a single failure cluster. These checkpoints (v0, v1, v2) represent intermediate refinements made by the Evolutionary Agent during offline diagnosis and repair, rather than separate outer-loop rollouts or additional VLA training. The base VLA policy remains frozen throughout.

Through case studies on representative tasks from RoboCasa and LIBERO-Pro, we demonstrate that:
\begin{enumerate}
\item \textbf{Stagnant Early Repairs:} Early revisions often yield marginal gains because they address local symptoms or overfit specific failure episodes, leading to performance plateaus.
\item \textbf{Bottleneck Identification:} A true ``Aha'' moment occurs when the agent isolates the critical physical state variable (e.g., grasp retention, end-effector re-alignment, or semantic approach geometry) required to restore the VLA's execution to its in-domain region.
\item \textbf{Discontinuous Scaling:} Resolving this root bottleneck produces a sharp, discontinuous increase in success rate, demonstrating that the harness acquires reusable embodied intelligence rather than task-specific trajectory tweaks.
\end{enumerate}

\subsubsection{``Aha'' Moments on LIBERO-Pro}
\begin{figure}[h]
\centering
\vspace{-5mm}
\includegraphics[width=0.95\linewidth]{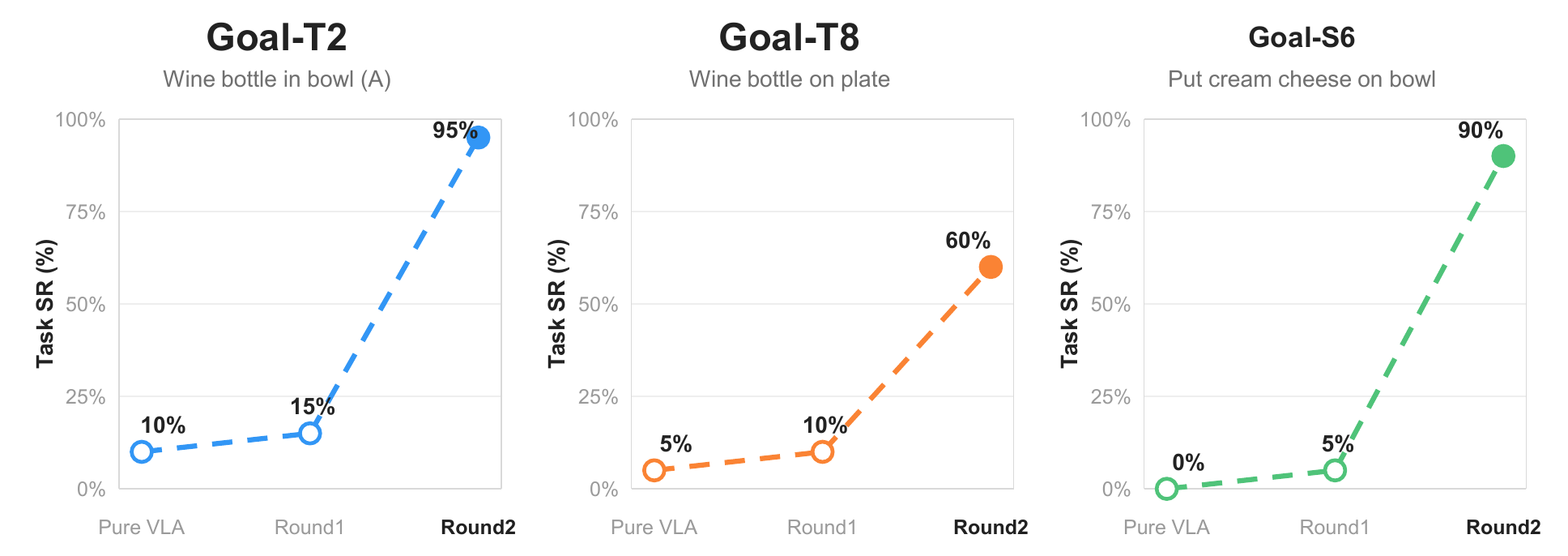}
\caption{\textbf{Physical-intelligence ``Aha'' moments on LIBERO-Pro.} Similar to RoboCasa, v0 denotes the Pure-VLA baseline and v1 represents early symptomatic repairs (e.g., staging or local gates) that yield stagnant performance. The ``Aha'' at v2 occurs when the agent identifies and resolves the decisive physical bottleneck (e.g., grasp retention or semantic approach), triggering a sharp increase in execution reliability.}
\label{fig:Libero-aha-case}
\end{figure}

Figure~\ref{fig:Libero-aha-case} shows representative ``Aha'' moments on LIBERO-Pro, where internal evolution progresses from symptomatic fixes to root-cause repairs. The checkpoints v0, v1, and v2 mirror the stagnant-to-breakthrough trajectory observed in RoboCasa.

For \textbf{Goal-T2} (placing a bottle in a bowl), the v1 revision introduces pre-grasp staging to improve approach geometry. However, performance remains stagnant ($10\% \rightarrow 15\%$) because it fails to address object loss during transport. The ``Aha'' occurs at v2, when the agent identifies \emph{grasp retention} as the decisive bottleneck. Implementing a retained-object critic that verifies stability before transport triggers a discontinuous jump to 95\% success.

A similar plateau appears in \textbf{Goal-T8}. The v1 revision improves success only marginally ($5\% \rightarrow 10\%$) as weak grasps during lifting remain unresolved. The v2 ``Aha'' moment involves implementing a recovery that monitors the full contact--grasp--retention sequence, ensuring stable retention before initiating transport. This raises success sharply to 60\%, as the system finally masters the physical transition from acquisition to transport.

In \textbf{Goal-S6} (placing cream cheese), the early v1 revision focuses on a release gate to prevent premature dropping. Because this only repairs a late-stage symptom, success stays stagnant at 5\%. The v2 ``Aha'' moment shifts the focus to the \emph{pre-contact phase}: the agent recognizes that the primary bottleneck is a lack of semantic progress during approach. By implementing a calibrated semantic pick-and-place recovery, the system resolves approach, grasp, and transport failures simultaneously, resulting in a sharp rise to 90\% success.

Across both benchmarks, these ``Aha'' moments confirm that the scalable unit of physical intelligence is the identification of state variables—such as grasp stability or approach geometry—that must be restored for reliable VLA execution.

\subsubsection{``Aha'' Moments on RoboCasa}
\begin{figure*}[h]
\centering
\vspace{-5mm}
\includegraphics[width=0.95\textwidth]{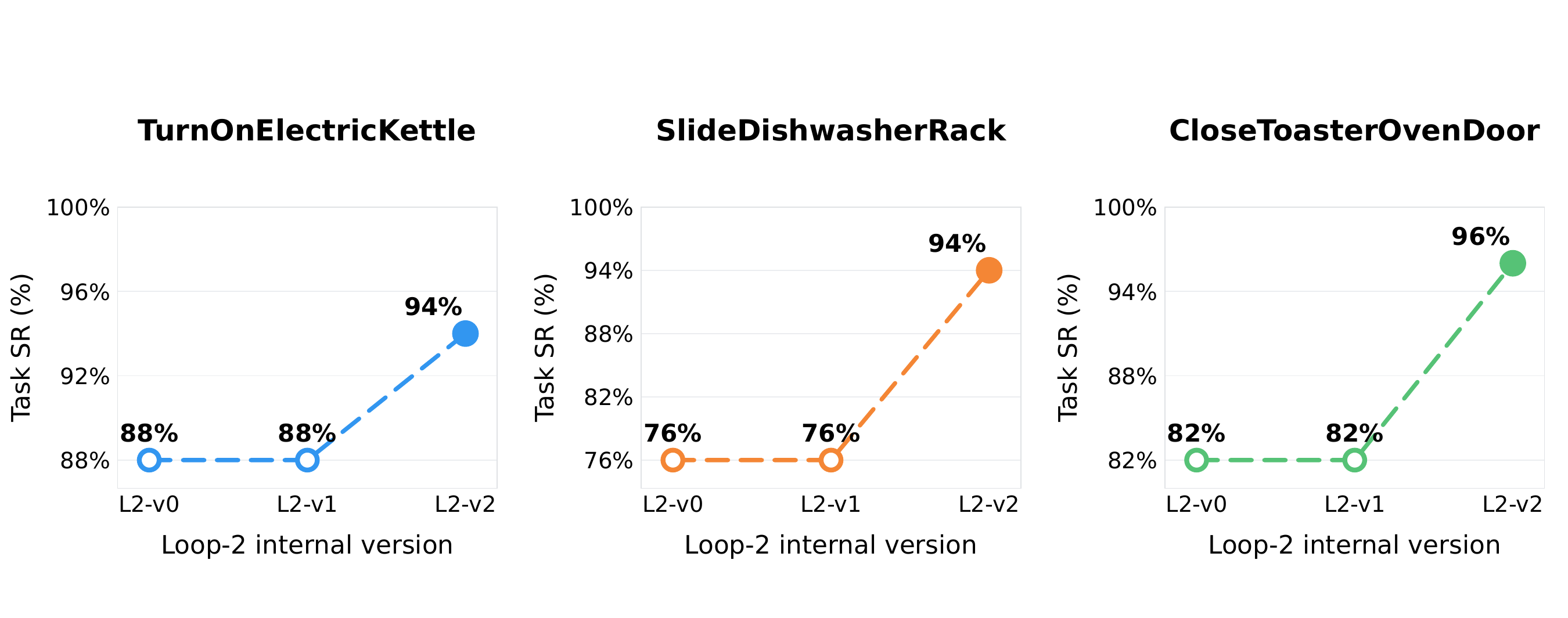}
\vspace{-5mm}
\caption{\textbf{Physical-intelligence ``Aha'' moments on RoboCasa.} L2-v0 is the original Pure-VLA baseline, while L2-v1 is an intermediate internal version. The flat L2-v0--L2-v1 segments summarize a period where early candidate fixes overfit individual failures. Once the agent identifies the key physical bottleneck (e.g., EEF alignment or centered contact), L2-v2 produces a sharp success-rate increase.}
\label{fig:robocasa-loop2-aha}
\end{figure*}

Figure~\ref{fig:robocasa-loop2-aha} illustrates this phenomenon on three representative RoboCasa tasks. We report success rates at three selected checkpoints within Loop-2 Stage-2. L2-v0 is the Pure-VLA baseline. L2-v1 represents an intermediate state where the agent is still analyzing failures and repeatedly revising the critic and recovery implementations. During this phase, most revisions produce only marginal gains or overfit observed failures, leaving success rates near the baseline.

An ``Aha'' moment occurs at L2-v2, where the internal version finally isolates the actual physical bottleneck. Success rates rise sharply: from 88\% to 94\% on \textsc{TurnOnElectricKettle}, 76\% to 94\% on \textsc{SlideDishwasherRack}, and 82\% to 96\% on \textsc{CloseToasterOvenDoor}. For \textsc{TurnOnElectricKettle}, the ``Aha'' was recognizing that simple EEF re-alignment restores the geometry expected by the VLA. For \textsc{SlideDishwasherRack}, the key was re-establishing centered contact after contact loss. These discontinuous gains reflect the harness's improved ability to restore the required physical state for the frozen VLA.

\subsection{Physical Intelligence Scaling and Zero-shot Capability}

We investigate the scaling of physical intelligence along two complementary axes: (1) \textbf{Intra-task scaling}, which measures how performance improves as a task accumulates Critic--Recovery mechanisms through reflection-driven evolution; and (2) \textbf{Cross-task zero-shot transfer}, which examines whether mechanisms discovered on a source task can be applied to unevolved target tasks sharing similar physical failure modes. Throughout all experiments, the underlying VLA policy remains frozen.

\subsubsection{Scaling and Zero-shot Capability on LIBERO-Pro}

\paragraph{Scaling through cumulative evolution.}
\begin{figure}[h]
    \centering
    \begin{subfigure}[t]{0.48\linewidth}
        \centering
        \includegraphics[width=\linewidth]{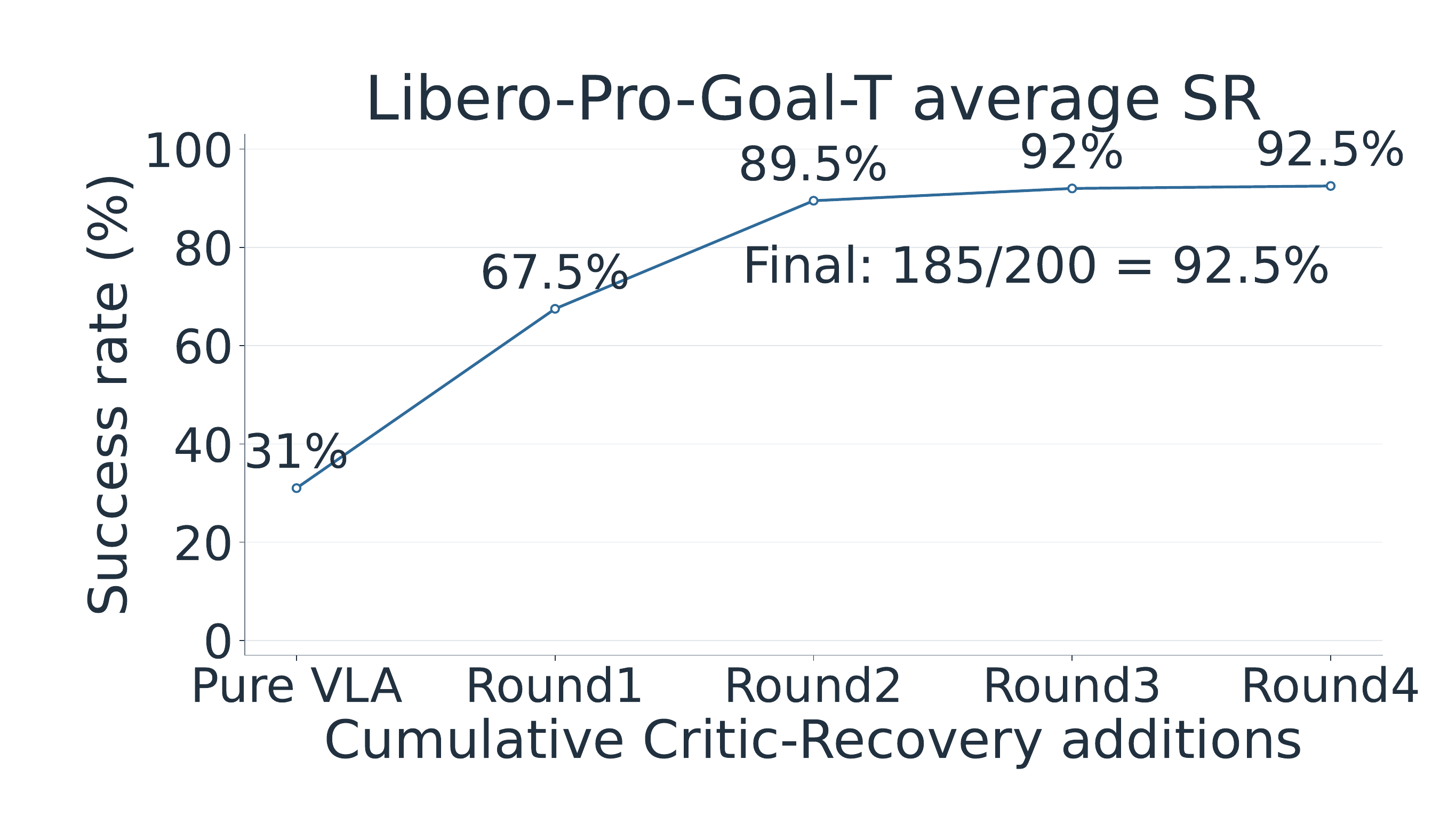}
        \caption{Task/instruction-redirection (T).}
        \label{fig:goal-t-average}
    \end{subfigure}
    \hfill
    \begin{subfigure}[t]{0.48\linewidth}
        \centering
        \includegraphics[width=\linewidth]{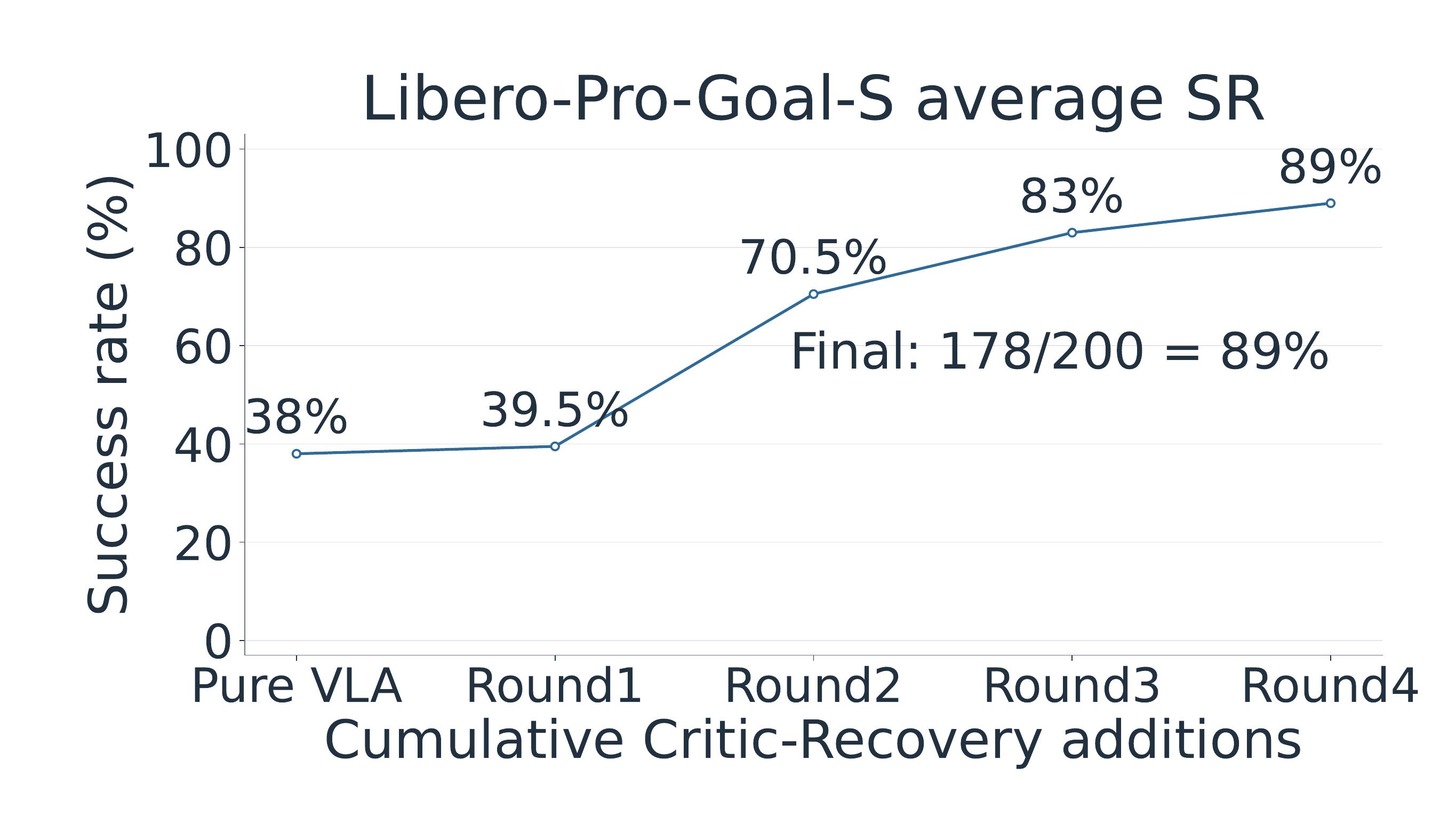}
        \caption{Swap/position-swap (S).}
        \label{fig:goal-s-average}
    \end{subfigure}
    \caption{\textbf{Cumulative physical-intelligence scaling on LIBERO-Pro
    Goal.} Average performance across ten tasks under Goal-T and Goal-S
    perturbations. Each point denotes a selected cumulative harness version.
    The base VLA is frozen throughout evolution.}
    \label{fig:Libero-aha-average}
\end{figure}

\begin{figure}[!h]
    \centering
    \includegraphics[width=0.75\linewidth]{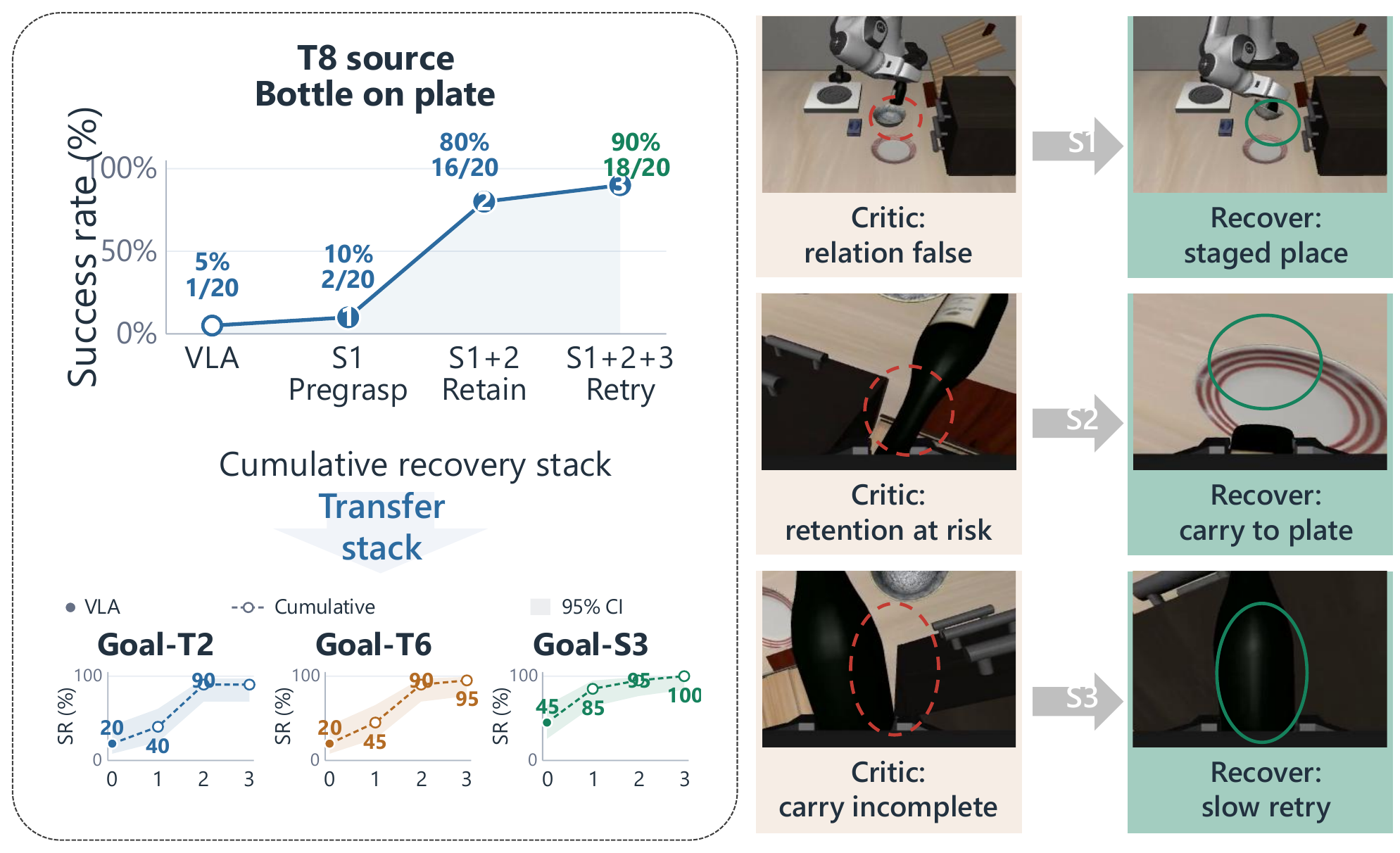}
    \caption{\textbf{Cross-task scaling from the Goal-T8 source task.}
    Three cumulative Critic--Recovery capabilities discovered on Goal-T8 are
    transferred to Goal-T2, Goal-T6, and Goal-S3. Curves report complete
    fixed-seed evaluations of each cumulative stack. Shaded regions denote
    Wilson $95\%$ confidence intervals. The right-hand frames illustrate the
    corresponding failure and recovery mechanisms rather than additional
    transfer measurements.}
    \label{fig:libero-t-transfer-scaling}
\end{figure}
Figure~\ref{fig:Libero-aha-average} reports the performance scaling on LIBERO-Pro Goal-T and Goal-S. As the harness accumulates Critic--Recovery mechanisms, the average success rate increases from $31.0\%$ to $92.5\%$ on Goal-T and from $38.0\%$ to $89.0\%$ on Goal-S. This improvement is achieved without increasing model capacity or fine-tuning weights. Instead, successive reflection rounds identify and repair recurring \textit{physical bottlenecks}, such as incorrect approach geometry, unstable grasp retention, and incomplete task relations. These results suggest that physical execution reliability can scale through the cumulative acquisition of compact runtime mechanisms.

\paragraph{Zero-shot Capability on Goal-T.}
We evaluate whether mechanisms discovered during evolution encode reusable physical principles. Figure~\ref{fig:libero-t-transfer-scaling} uses Goal-T8 (placing a wine bottle) as the source task. Three cumulative capabilities were discovered: (1) pre-grasp staging, (2) a grasp-retention Critic, and (3) a failure-gated retry. When applied \textit{zero-shot} to Goal-T2, Goal-T6, and Goal-S3, these mechanisms yield significant performance gains (e.g., $9/20 \rightarrow 20/20$ on Goal-S3). The transfer succeeds because the mechanisms operate on \textit{task-independent physical variables}---such as EEF alignment and contact stability---rather than memorizing source-task trajectories.


\begin{figure}[h]
    \centering
    \includegraphics[width=0.75\linewidth]{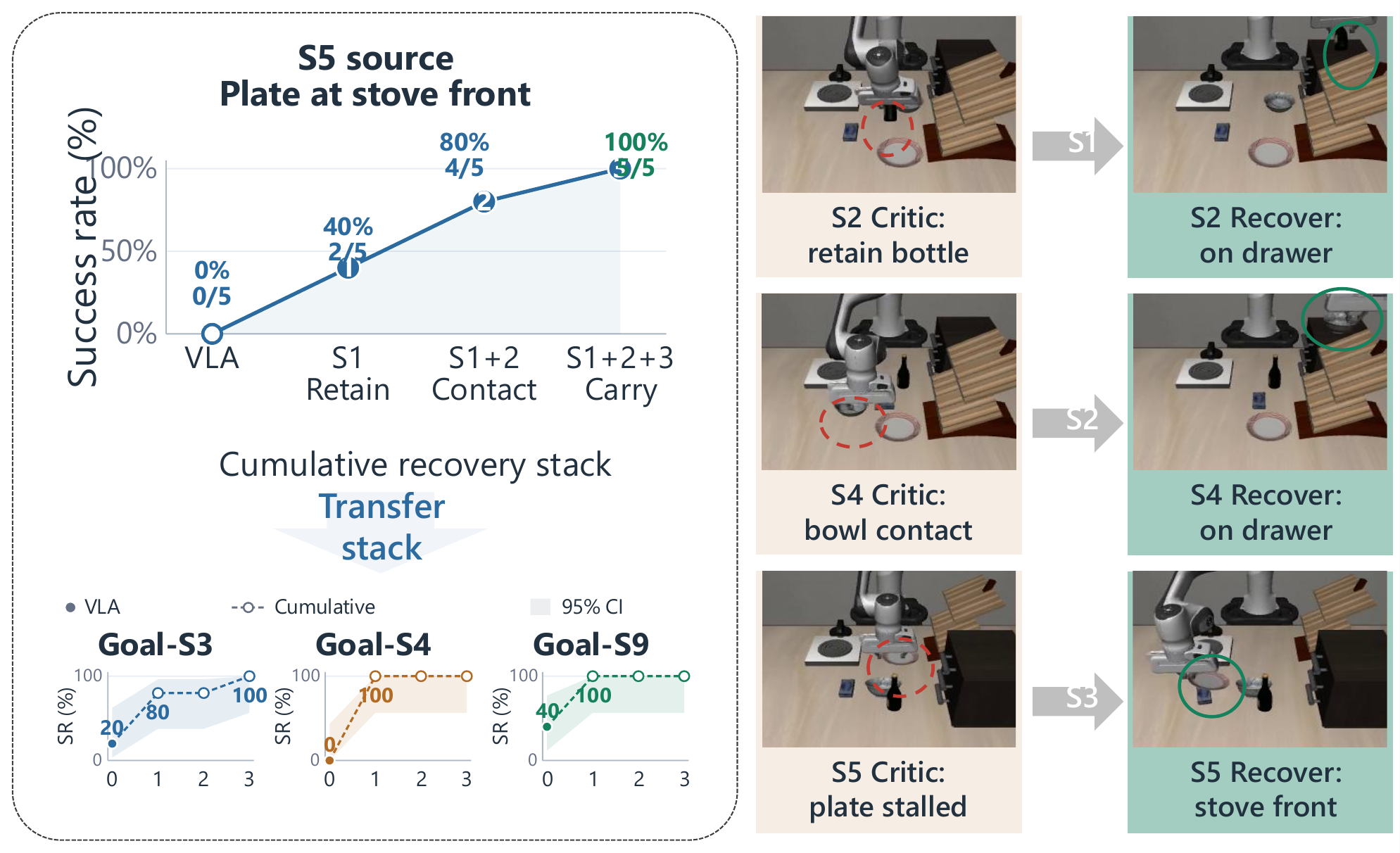}
    \caption{\textbf{Cross-task scaling from the Goal-S5 source task.}
    Cumulative Critic--Recovery capabilities discovered on Goal-S5 are
    transferred to Goal-S3, Goal-S4, and Goal-S9. Within each task, all four
    arms use the same fixed seeds, policy RNGs, checkpoint, and execution
    budget. Shaded regions denote Wilson $95\%$ confidence intervals.}
    \label{fig:libero-s-transfer-scaling}
    \vspace{-5mm}
\end{figure}
\paragraph{Zero-shot Capability on Goal-S.}
\begin{wrapfigure}{r}{0.45\textwidth}
    \centering
    \includegraphics[width=0.45\textwidth]
    {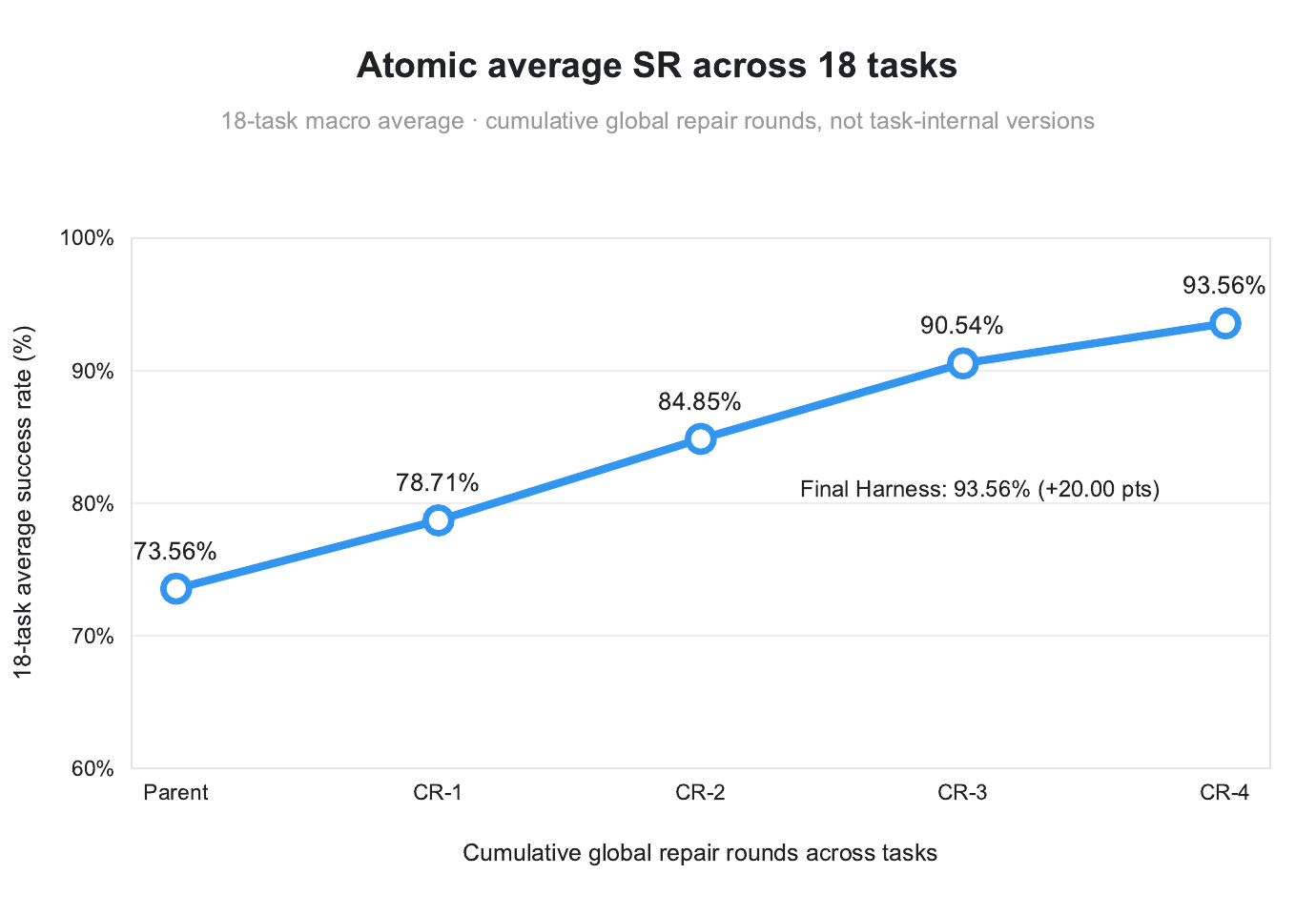}
    \caption{\textbf{Cumulative reflection scaling across 18 RoboCasa tasks.}
    The macro-average success rate increases from 73.56\% for the frozen
    parent harness to 78.71\%, 84.85\%, 90.54\%, and 93.56\% after four
    cumulative global repair rounds. Each checkpoint retains the previously
    validated critic, recovery, and tool capabilities.}
    \label{fig:robocasa-atomic-scaling}
    \vspace{-30pt}
\end{wrapfigure}
Figure~\ref{fig:libero-s-transfer-scaling} provides a complementary study using Goal-S5 as the source task. The evolved mechanisms include contact-qualified takeovers based on EEF--object proximity and a stalled-command carry gate for robust transport. Without changing the frozen policy, the same cumulative stack was evaluated \textit{zero-shot} on Goal-S3, Goal-S4, and Goal-S9. The results show that mechanisms learned from a single task can successfully repair shared grasp, contact, and transport failures across multiple unevolved tasks, demonstrating that the scalable unit is a reusable mapping from physical failure states to recovery behaviors.

\subsubsection{Scaling and Zero-shot Capability on RoboCasa}

\paragraph{Scaling through cumulative evolution.}
The aggregate scaling trend on 18 RoboCasa tasks is shown in Figure~\ref{fig:robocasa-atomic-scaling}. Through four rounds of global reflection and repair, the macro-average success rate increases from $73.56\%$ to $93.56\%$. Since the VLA is frozen, this $20.00$ percentage-point gain reflects the accumulation of validated execution knowledge within the harness. The scaling variable here is the cumulative reflection experience, which continuously improves the capability of a fixed policy by addressing deeper physical bottlenecks.

\paragraph{Zero-shot transfer of Pick-and-Place capabilities.}

\begin{figure*}[h]
    \centering
    \includegraphics[width=0.95\textwidth]
    {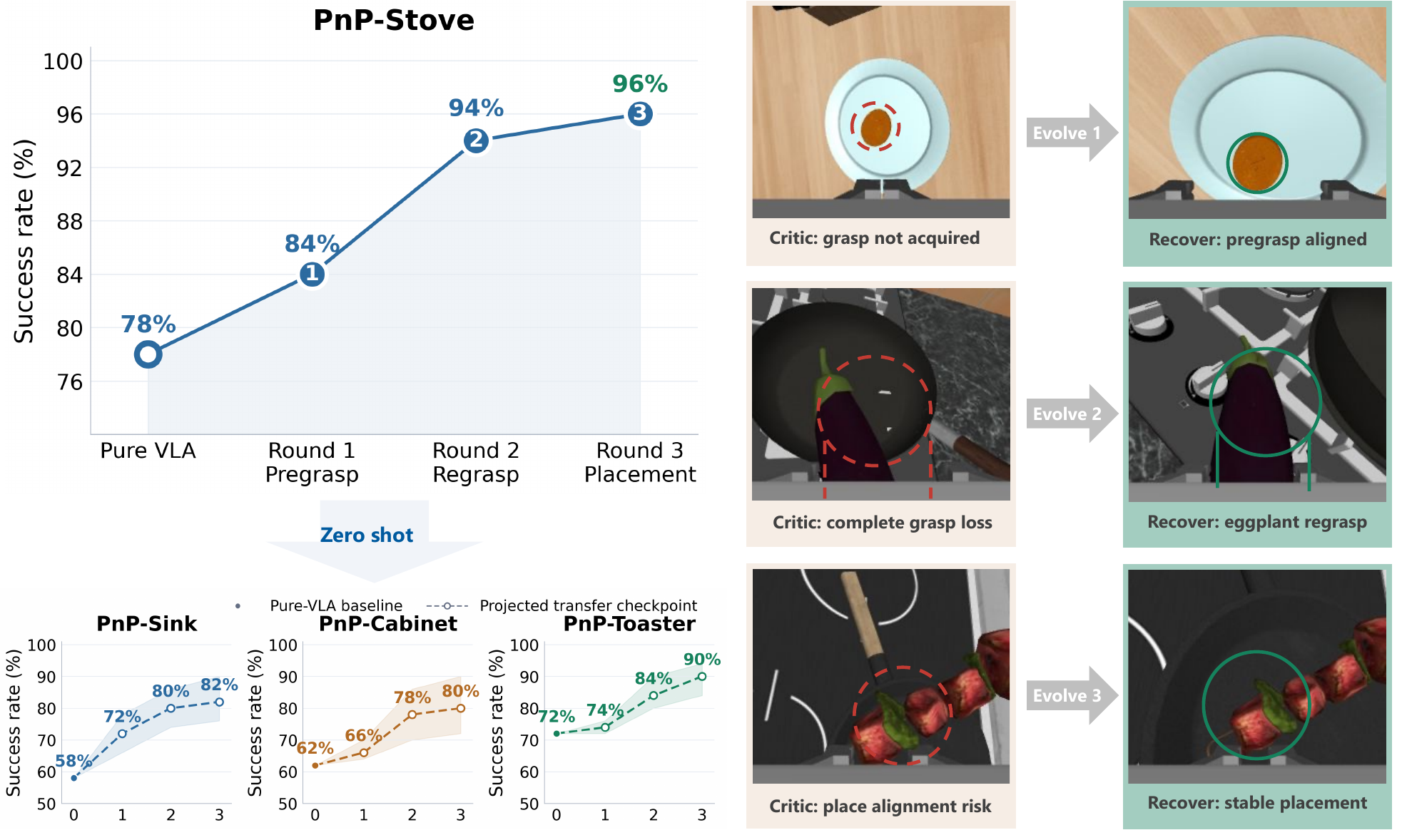}
    \caption{\textbf{Reflection-driven scaling and transfer on PnP tasks.}
    On PnP-Stove, successive rounds add object-relative pregrasp alignment,
    bounded regrasp after grasp loss, and stable placement. The lower panels
    apply the cumulative checkpoints to PnP-Sink, PnP-Cabinet, and PnP-Toaster
    without an additional training or evolution loop. The right-hand panels
    show the failure signatures identified by the critic and the corresponding
    recoveries.}
    \label{fig:robocasa-pnp-scaling}
\end{figure*}

We examine the zero-shot transferability of mechanisms discovered on \textsc{PnP-Stove} (Figure~\ref{fig:robocasa-pnp-scaling}). Three cumulative rounds on the source task produced: (1) \textit{pre-grasp alignment}, (2) \textit{re-grasp after grasp loss}, and (3) \textit{stable placement}. Without any additional training or evolution loops, we apply this cumulative stack to \textsc{PnP-Sink}, \textsc{PnP-Cabinet}, and \textsc{PnP-Toaster}. The macro-average success rate over these transfer tasks increases from $64\%$ to $84\%$. This $20$ percentage-point zero-shot gain demonstrates that the agent learns general pick-and-place principles that transcend specific object identities or furniture geometries.

\paragraph{Zero-shot transfer of Articulated interaction capabilities.}

\begin{figure*}[h]
    \centering
    \includegraphics[width=0.95\textwidth]
    {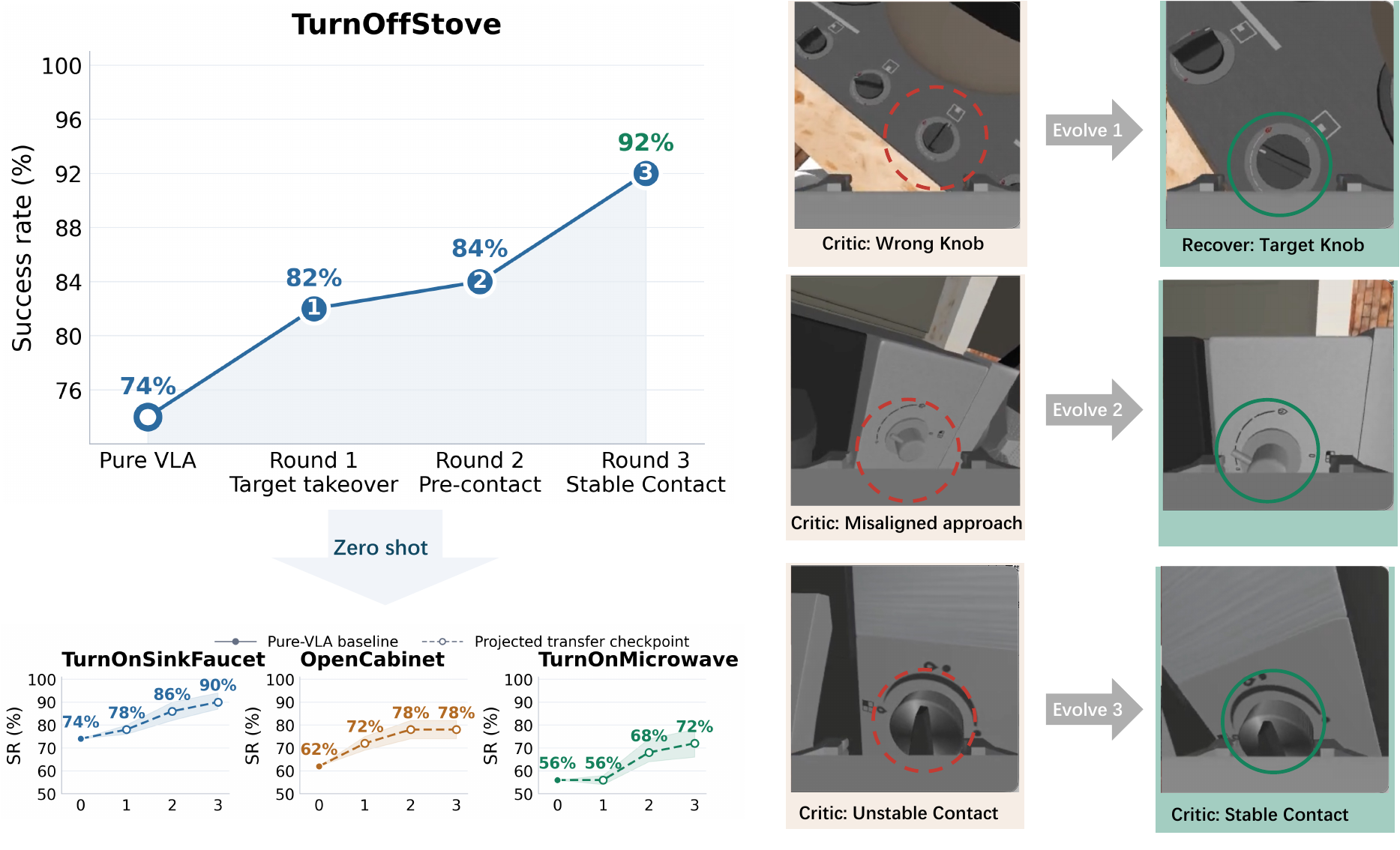}
    \caption{\textbf{Reflection-driven scaling and transfer on articulated
    interaction tasks.} On TurnOffStove, successive rounds add target
    localization, collision-aware pre-contact approach, and stable
    EEF--target contact. The lower panels apply the cumulative checkpoints to
    TurnOnSinkFaucet, OpenCabinet, and TurnOnMicrowave without an additional
    training or evolution loop. The right-hand panels show the diagnosed
    target, approach, and contact failures together with the recovered states.}
    \label{fig:robocasa-turnoff-scaling}
\end{figure*}

A second study focuses on \textsc{TurnOffStove}, where the agent evolves (1) \textit{target localization}, (2) \textit{collision-aware approach}, and (3) \textit{stable-contact} skills (Figure~\ref{fig:robocasa-turnoff-scaling}). These skills describe physical invariants shared by many articulated-interaction tasks. We evaluate these source-task mechanisms \textit{zero-shot} on \textsc{TurnOnSinkFaucet}, \textsc{OpenCabinet}, and \textsc{TurnOnMicrowave}. The transfer-task macro-average increases from $64\%$ to $80\%$, a $16$ percentage-point improvement without target-task adaptation. These results confirm that the scalable unit in our framework is not a task-specific trajectory, but a reusable mapping from observable physical failure states to robust recovery behaviors.

\subsection{Results on Simulation benchmark}
\paragraph{RoboCasa}
Table~\ref{tab:robocasa-atomic} reports the success rate on all 18
RoboCasa Atomic-Seen tasks. We use compact task identifiers in the main text;
their official RoboCasa names are listed in
Appendix~\ref{app:atomic-task-mapping}. Compared with the frozen GR00T
post-trained VLA, \sys improves the macro-average success rate from 73.56\%
to 93.56\%, an absolute gain of 20.00 percentage points. The improvement is
consistent across all tasks and is especially large on contact-rich or
long-horizon manipulation tasks such as T4, T5, and T15.
\begin{table}[h]
    \centering
    \caption{\textbf{Success rates (\%) on 18 RoboCasa Atomic-Seen tasks.}
    Task identifiers follow Appendix~\ref{app:atomic-task-mapping}. The best
    result for each task is shown in bold. ``Avg.'' is the macro-average over
    all 18 tasks.}
    \label{tab:robocasa-atomic}
    \small
    \setlength{\tabcolsep}{3.2pt}
    \renewcommand{\arraystretch}{1.08}
    \resizebox{0.7\columnwidth}{!}{%
    \begin{tabular}{lcccccccccc}
        \toprule
        \textbf{Method} & \textbf{T1} & \textbf{T2} & \textbf{T3}
        & \textbf{T4} & \textbf{T5} & \textbf{T6} & \textbf{T7}
        & \textbf{T8} & \textbf{T9} & \textbf{Avg.} \\
        \midrule
        Pure VLA (GR00T)
        & 78 & 74 & 78 & 58 & 48 & 62 & 72 & 74 & 70 & 73.56 \\
        \sys
        & \textbf{96} & \textbf{92} & \textbf{94} & \textbf{96}
        & \textbf{86} & \textbf{80} & \textbf{96} & \textbf{96}
        & \textbf{86} & \textbf{93.56} \\
        \midrule
        \textbf{Method} & \textbf{T10} & \textbf{T11} & \textbf{T12}
        & \textbf{T13} & \textbf{T14} & \textbf{T15} & \textbf{T16}
        & \textbf{T17} & \textbf{T18} & \textbf{Avg.} \\
        \midrule
        Pure VLA (GR00T)
        & 62 & 92 & 76 & 88 & 96 & 50 & 90 & 82 & 74 & 73.56 \\
        \sys
        & \textbf{94} & \textbf{98} & \textbf{94} & \textbf{94}
        & \textbf{100} & \textbf{100} & \textbf{94} & \textbf{96}
        & \textbf{92} & \textbf{93.56} \\
        \bottomrule
    \end{tabular}%
    }
\end{table}

\paragraph{Libero-Pro}
Table~\ref{tab:libero-pro} reports results on 40 task-setting pairs across
the Goal and LIBERO-10 suites. Compared with the frozen $\pi_{0.5}$ baseline,
\sys improves the overall macro-average from 32.00\% to 71.13\%, an absolute
gain of 39.13 percentage points. The gains are largest on Goal (T) and Goal
(S), whose averages increase from 31.0\% to 92.5\% and from 38.0\% to 89.0\%,
respectively. On LIBERO-10, \sys improves the T and S settings from 50.0\% to
63.0\% and from 9.0\% to 40.0\%. Overall, \sys improves 32 task-setting pairs
and matches the baseline on the remaining eight.

\begin{table}[h]
\centering
\caption{\textbf{Success rates (\%) on LIBERO-Pro.} The best result for each
task is shown in bold. ``Average'' is the macro-average over 10 tasks.}
\label{tab:libero-pro}
\setlength{\tabcolsep}{5pt}
\renewcommand{\arraystretch}{1.08}
\resizebox{\linewidth}{!}{
\begin{tabular}{llccccccccccc}
\toprule
\textbf{Setting} & \textbf{Method}
& \textbf{Task 0} & \textbf{Task 1} & \textbf{Task 2} & \textbf{Task 3}
& \textbf{Task 4} & \textbf{Task 5} & \textbf{Task 6} & \textbf{Task 7}
& \textbf{Task 8} & \textbf{Task 9} & \textbf{Average} \\
\midrule

\multirow{2}{*}{\textbf{Goal (T)}}
& $\pi_{0.5}$
& 0.0 & 95.0 & 10.0 & 0.0 & \textbf{100.0} & 0.0 & 20.0 & 80.0 & 5.0 & 0.0 & 31.0 \\
& \sys
& \textbf{80.0} & \textbf{100.0} & \textbf{95.0} & \textbf{80.0}
& \textbf{100.0} & \textbf{100.0} & \textbf{95.0} & \textbf{95.0}
& \textbf{80.0} & \textbf{100.0} & \textbf{92.5} \\
\midrule

\multirow{2}{*}{\textbf{Goal (S)}}
& $\pi_{0.5}$
& 0.0 & 60.0 & 0.0 & 45.0 & 0.0 & 0.0 & 0.0
& \textbf{100.0} & \textbf{100.0} & 75.0 & 38.0 \\
& \sys
& \textbf{90.0} & \textbf{65.0} & \textbf{80.0} & \textbf{85.0}
& \textbf{95.0} & \textbf{95.0} & \textbf{100.0}
& \textbf{100.0} & \textbf{100.0} & \textbf{80.0} & \textbf{89.0} \\
\midrule

\multirow{2}{*}{\textbf{LIBERO-10 (T)}}
& $\pi_{0.5}$
& 5.0 & \textbf{95.0} & 95.0 & \textbf{0.0} & 0.0
& 80.0 & 85.0 & 75.0 & 65.0 & \textbf{0.0} & 50.0 \\
& \sys
& \textbf{35.0} & \textbf{95.0} & \textbf{100.0} & \textbf{0.0}
& \textbf{25.0} & \textbf{100.0} & \textbf{95.0} & \textbf{80.0}
& \textbf{100.0} & \textbf{0.0} & \textbf{63.0} \\
\midrule

\multirow{2}{*}{\textbf{LIBERO-10 (S)}}
& $\pi_{0.5}$
& 0.0 & 35.0 & 0.0 & 0.0 & 5.0 & 50.0 & 0.0
& \textbf{0.0} & \textbf{0.0} & 0.0 & 9.0 \\
& \sys
& \textbf{90.0} & \textbf{50.0} & \textbf{95.0} & \textbf{75.0}
& \textbf{15.0} & \textbf{65.0} & \textbf{5.0}
& \textbf{0.0} & \textbf{0.0} & \textbf{5.0} & \textbf{40.0} \\

\bottomrule
\end{tabular}
}
\end{table}

\paragraph{Final best accuracy vs other SOTA.}

On LIBERO-Pro, the final \sys configuration achieves 92.5\% and 89.0\%
average success on Goal (T) and Goal (S), and 63.0\% and 40.0\% on the
corresponding LIBERO-10 settings. Compared with the strong frozen
$\pi_{0.5}$ VLA, this raises the overall average from 32.00\% to 71.13\%.
Notably, these gains require no additional VLA training: \sys improves or
matches the baseline on every task-setting pair by detecting execution
failures and applying targeted recovery at test time.

\subsection{Case Studies}
\begin{figure*}[h]
    \centering
    \includegraphics[width=0.8\textwidth]{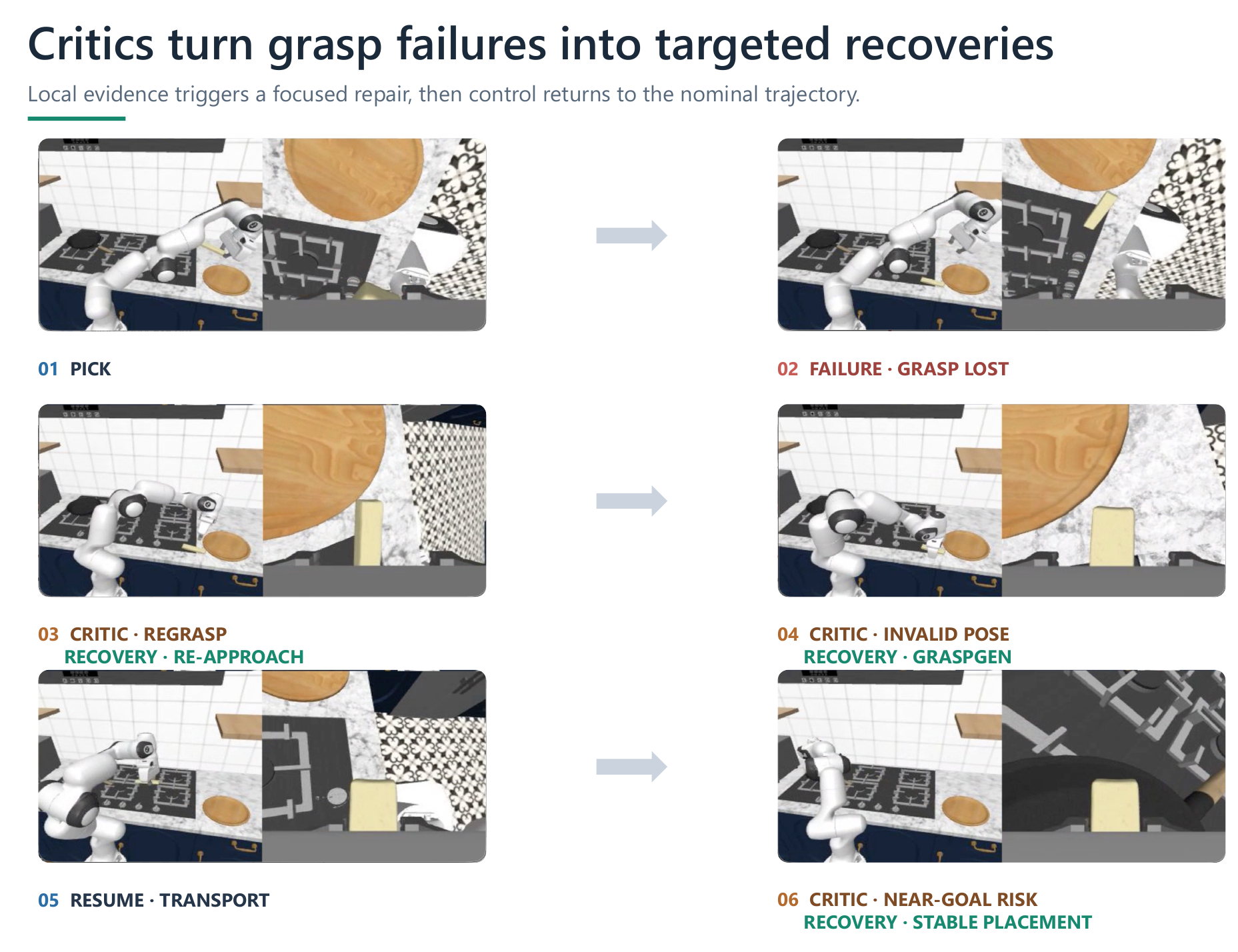}
    \caption{\textbf{A sequence of critic--recovery interventions during aPnP episode.} The VLA first picks and transports the object.  A critic detects a lost grasp and triggers a re-approach recovery; an invalid
    re-grasp pose then triggers \textsc{GraspGen} to synthesize a feasible
    grasp pose; near the goal, a final critic invokes stable \textsc{CAP}
    placement.  After each local repair, control returns to the nominal VLA
    only after the recovery state is verified.}
    \label{fig:pnp-transport-case}
\end{figure*}
\paragraph{Repeated critic--recovery interventions in Robocasa}
Figure~\ref{fig:pnp-transport-case} illustrates a representative PnP
execution in which the harness resolves several distinct failures within a
single episode.  The base VLA first completes the pick and begins transporting
the object.  When the object slips from the gripper, the runtime critic detects
the loss of grasp and interrupts the nominal trajectory.  The corresponding
recovery executes a controlled re-approach to the object and establishes a new
grasp, after which control returns to the VLA for transport.

The second intervention is triggered when the initial re-grasp pose is
infeasible.  Rather than repeatedly issuing the same motion, the critic reports
the invalid approach geometry and the recovery invokes \textsc{GraspGen} to
generate a feasible end-effector pose.  The generated pose is then executed to
re-grasp the object, restoring the contact configuration required for stable
transport.  Finally, as the object approaches the destination, the critic
detects near-goal placement risk and hands control to a stable placement
recovery based on \textsc{CAP}.  This recovery completes the final placement
before the harness verifies the re-entry conditions and terminates the episode.

This example demonstrates that the harness is not limited to a single
episode-level retry: critics can monitor the execution continuously, trigger
different recoveries for different physical failure modes, and return control
to the VLA after each local repair.  The resulting behavior is a sequence of
targeted interventions---re-approach, GraspGen-based re-grasp, and CAP-based
placement---that preserves the nominal policy while making the long-horizon
execution robust to contact and grasp failures.

\begin{figure}[h]
    \centering
    \includegraphics[width=0.8\linewidth]{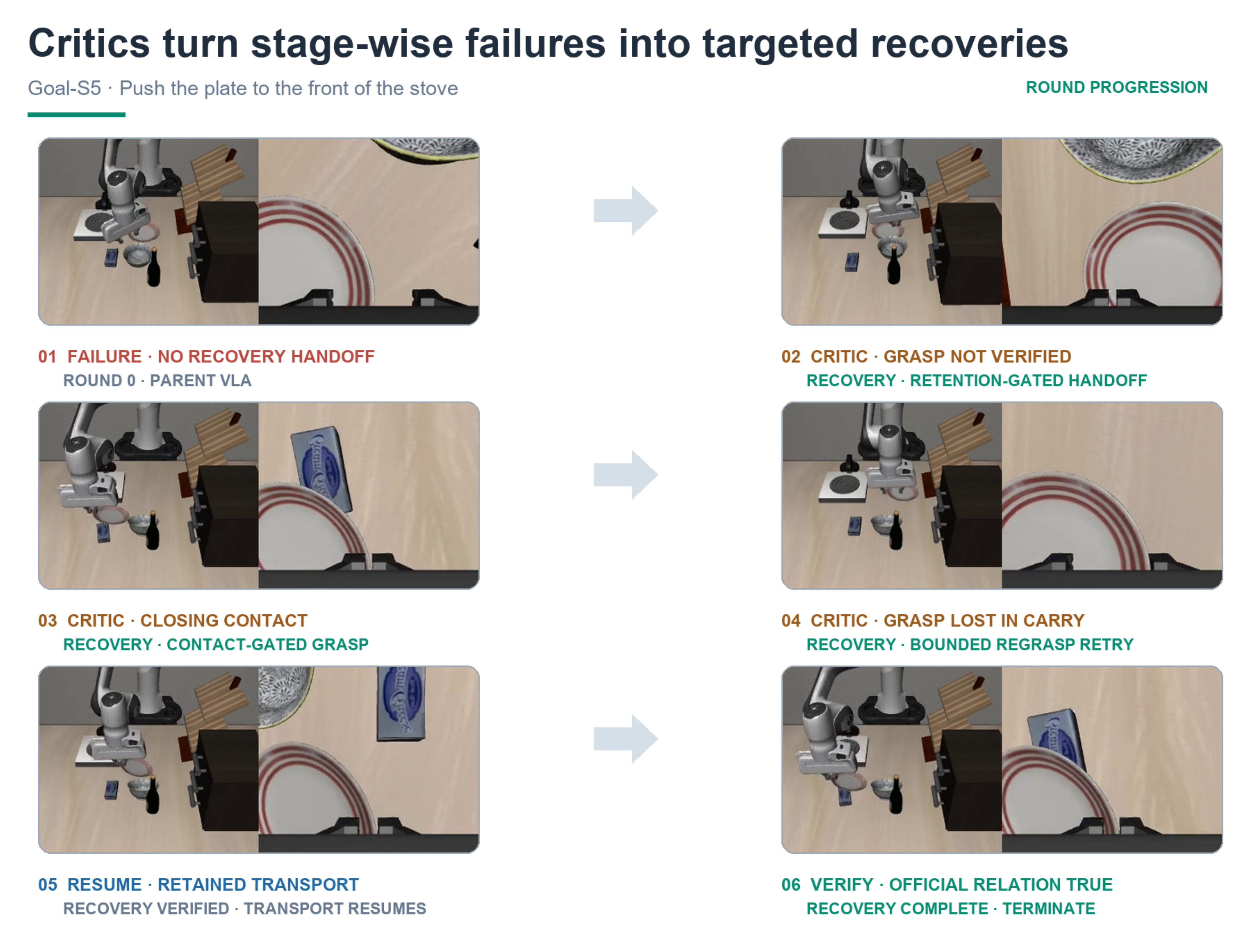}
    \caption{\textbf{Critic--recovery promotion progressively moves the
    failure frontier in LIBERO-Pro Goal-S5.} The parent VLA exhausts its
    execution budget without a timely recovery handoff. Successive promotion
    rounds introduce a retention-gated handoff, a contact-gated grasp
    recovery, and a bounded re-grasp retry. Each addition resolves the
    previously exposed failure and reveals a later-stage bottleneck, until
    retained transport resumes and the official stove-front relation is
    satisfied.}
    \label{fig:libero-goal-s5-case}
\end{figure}

\paragraph{Moving the failure frontier across promotion rounds in Libero-Pro}
Figure~\ref{fig:libero-goal-s5-case} presents the corresponding iterative
behavior in LIBERO-Pro Goal-S5, where the robot must push the plate to the
front of the stove. Unlike the RoboCasa example, the panels do not depict
multiple independent critic activations within one episode. Instead, they
summarize representative episodes from successive promotion rounds and show
how each promoted critic--recovery skill moves the earliest unresolved failure
to a later execution stage.

In Round~0, the parent VLA continues executing without a timely handoff to a
bounded recovery and eventually exhausts the episode budget. Round~1
introduces a retention-gated handoff. This resolves the absence of intervention,
but the episode now stops because a stable grasp cannot be verified. Round~2
adds a closing-contact critic and a contact-gated grasp recovery. The plate is
successfully acquired, exposing a new failure during transport: the retained
grasp is lost while carrying the plate. Thus, the failure frontier has moved
from recovery invocation, through grasp verification, to carry retention.

Round~3 addresses this later-stage failure with a bounded re-grasp retry. When
the critic detects loss of the plate during carry, the recovery does not restart
the entire episode or repeat the nominal action indefinitely. It performs a
bounded re-grasp attempt, verifies the restored retention state, and resumes
transport toward the stove-front region. The episode terminates only after the
official LIBERO relation becomes true. Importantly, Rounds~2 and~3 use the same
environment seed and policy RNG, directly isolating the contribution of the
carry-retry mechanism. Separate same-seed adjacent-round comparisons similarly
validate the retention-handoff and contact-gated grasp additions.

Together, the two case studies expose complementary forms of compositionality.
RoboCasa composes several specialized recoveries online within a single
long-horizon episode, whereas LIBERO-Pro composes promoted critic--recovery
skills across outer-loop iterations. In both cases, improvement arises from
localizing the earliest unresolved failure, applying a bounded repair, and
verifying the repaired state before allowing execution to advance.

\subsection{Z-Infra Performance Evaluation}

Z-Infra enables effective throughput scaling under high concurrency with controlled latency growth, achieving 7.7$\times$ higher throughput than \textbf{Ours without Z-Infra} and 12.8$\times$ higher than \textbf{RPent}~\cite{zhang2026harnessvla} under moderate concurrency, while reducing per-episode latency by 11.9$\times$ compared to RPent.
We evaluate throughput scaling behavior across different concurrency levels (1, 8, 16, 32, 64) on LIBERO Goal~\cite{liu2023libero}.
All experiments run on 8$\times$A100 GPUs with identical model checkpoints and action budgets.

\begin{figure}
    \centering
    \includegraphics[width=0.65\linewidth]{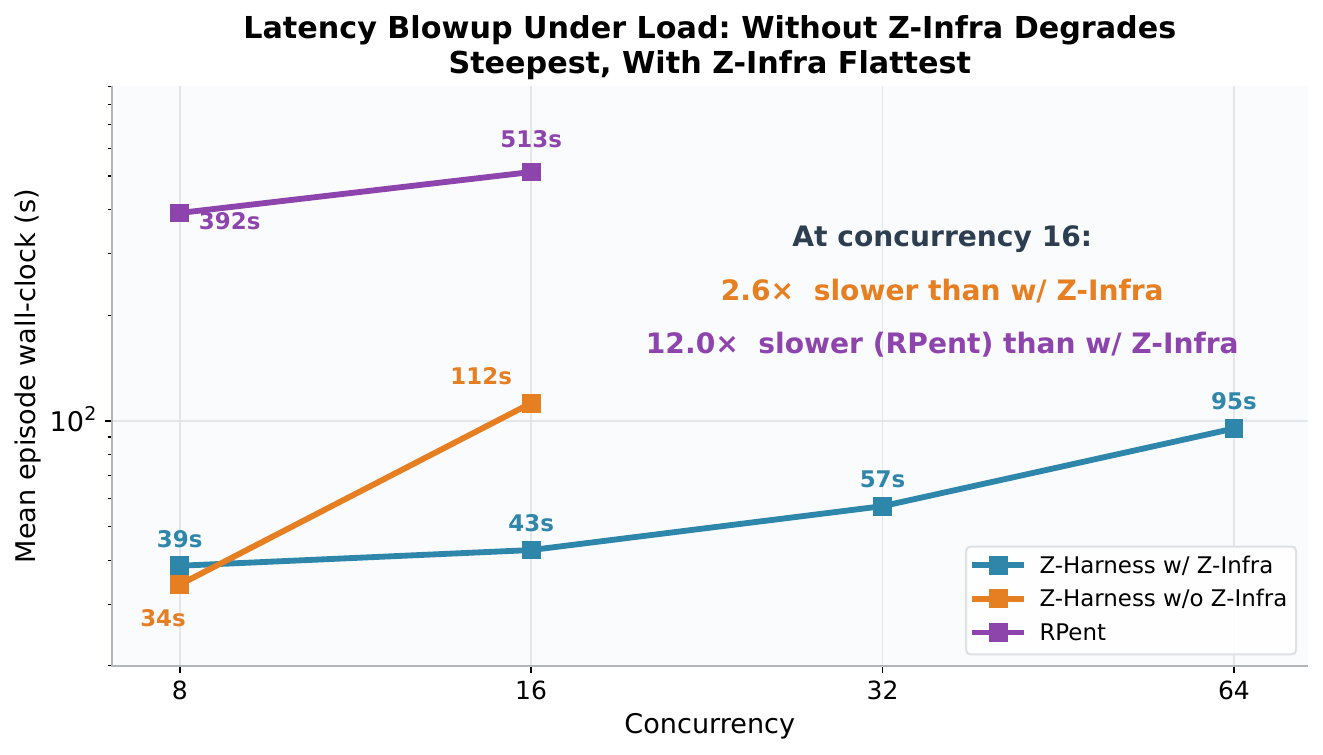}
    \caption{Per-episode latency versus concurrency. Z-Infra maintains controlled growth while baselines exhibit explosive degradation or inherently high overhead. Both baselines encounter OOM beyond concurrency 16.}
    \label{fig:rollout-latency}
\end{figure}

\begin{figure}
    \centering
    \includegraphics[width=0.65\linewidth]{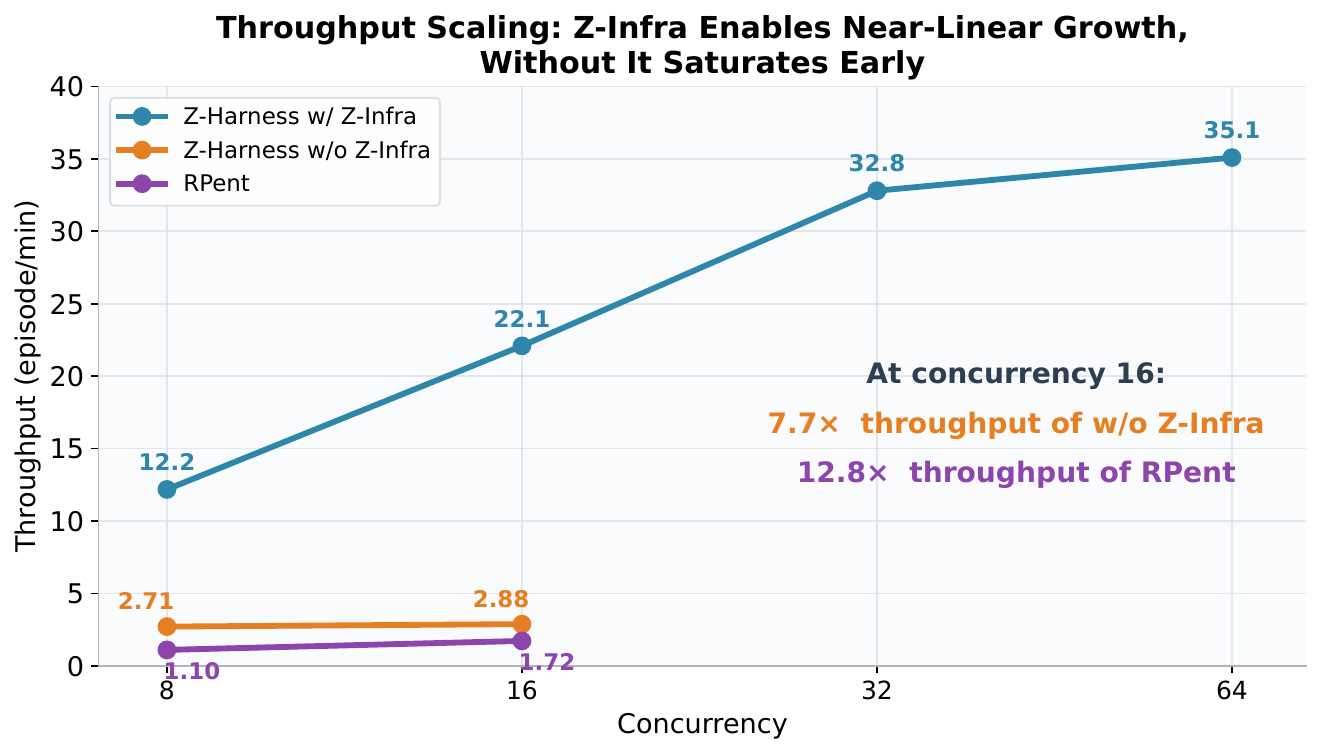}
    \caption{Rollout throughput versus concurrency. Z-Infra scales to 35.1 ep/min at concurrency 64, achieving 7.7× and 12.8× higher throughput than baselines at concurrency 16. Baselines fail beyond concurrency 16 due to OOM.}
    \label{fig:rollout-throughput}
\end{figure}

\paragraph{Latency Under Load.} Z-Infra maintains controlled latency growth under high concurrency, while baselines either exhibit explosive degradation or inherently high overhead. 
As shown in Figure~\ref{fig:rollout-latency}, Z-Infra's per-episode latency increases from 39s at concurrency 8 to 57s at concurrency 32 (46\% increase), then to 95s at concurrency 64. 
In contrast, Ours w/o Z-Infra experiences explosive latency degradation—from 34s at concurrency 8 to 112s at concurrency 16 (3.3$\times$ increase). 
RPent exhibits inherently high latency (392s$\to$513s), as agent-in-the-loop design requires LLM API calls at every decision point. 
Our approach reduces agent overhead by invoking agents only during offline Reflection \& Evolve phases, while online rollouts execute pure VLA policy under lightweight runtime critics. 
Z-Infra further ensures that latency growth remains sublinear even as concurrency scales, maintaining efficient resource utilization without cascading delays.
\paragraph{Throughput Scaling.} Z-Infra achieves effective throughput scaling up to hardware saturation, while baselines fail to scale beyond low concurrency levels due to resource contention. 
As shown in Figure~\ref{fig:rollout-throughput}, Z-Infra's throughput increases from 12.18 ep/min at concurrency 8 to 32.8 ep/min at concurrency 32 (2.7$\times$ speedup), reaching 22.09 ep/min at concurrency 16—7.7$\times$ higher than Ours w/o Z-Infra (2.88 ep/min) and 12.8$\times$ higher than RPent (1.72 ep/min). 
Beyond concurrency 32, throughput plateaus at 35.1 ep/min (concurrency 64), indicating saturation of the 8-GPU configuration. 
The performance gap stems from Z-Infra's persistent worker design with dynamic batching and asynchronous scheduling, which eliminates per-episode initialization overhead and maintains high GPU utilization under variable request patterns.

%% file: tex/related.tex
\section{Related Work}
We organize related work around the motivation for closed-loop embodied learning. First, physical intelligence is scaling through both foundation models and agent harnesses, but the deployment gap remains. Second, self-evolving agents have shown strong progress in digital domains, yet embodied self-evolution is harder because the agent must interpret and act in a continuous physical world. Third, existing rollout systems are mostly designed for standard RL workloads, while embodied self-evolution requires heterogeneous model, tool, simulator, robot, memory, and critic execution.

\subsection{Embodied Foundation Model Gaps for Physical Intelligence Scaling}
Recent physical intelligence research is advancing along two complementary paths. The first path trains end-to-end policy models for general robot control, including VLA models such as RT-1/RT-2, OpenVLA, $\pi_0$, $\pi_{0.5}$, GR00T, CogACT, UniVLA, and FAST, as well as world-action models such as DreamZero, Cosmos-Policy, and FAST-WAM \citep{brohan2022rt1,brohan2023rt2,kim2024openvla,black2024pi0,black2025pi05,bjorck2025groot,li2024cogact,bu2025univla,pertsch2025fast,dreamzero2026,cosmospolicy2026,fastwam2026}. This line is promising, but real deployment still exposes a \textbf{persistent gap}: data are expensive, physical distributions shift, and small execution errors can cascade in long-horizon tasks, which limits end-to-end models from moving beyond demonstrations into real-world productivity. 

The second path explores embodied agent capabilities by using LLMs, code, tools, memory, planning, verification, and recovery around foundation policies to fill this gap. Earlier systems such as PaLM-E and Code as Policies demonstrated the potential of language models for embodied reasoning and planning, while RoboCat showed that an agent can broaden its manipulation competence by collecting data from its own attempts across tasks and embodiments \citep{driess2023palme,liang2023codeaspolicies,bousmalis2023robocat}. Recent systems and demonstrations, including Claude Plays Robotics, CaP-X, HarnessVLA, and Guava, further suggest that coding agents and execution scaffolds can make frozen or pretrained robot policies more reliable without waiting for a fully solved end-to-end policy \citep{anthropic2026clauderobotics,capx2026,zhang2026harnessvla,liu2026guava}. The state of the art is therefore shifting from relying only on a single end-to-end policy toward scaffolded systems that coordinate policies with planning, memory, tools, verification, and recovery. 

However, many embodied agents remain episodic: they may recover within a trial, but they rarely convert execution traces into governed long-term improvements. Our work follows the embodied-agent path by studying a closed-loop harness that turns deployment feedback into reusable critics, recoveries, and future rollouts.

\subsection{Live Harness Gaps for Embodied Self-Evolution}
General agent self-evolution has mainly been studied in digital domains where execution is cheap, interfaces are discrete, APIs are explicit, logs are exact, and evaluation is relatively easy. Reflexion, Self-Refine, Voyager, and OPRO show that language or game agents can improve through verbal feedback, iterative revision, executable skill libraries, or search over prompts and programs \citep{shinn2023reflexion,madaan2023selfrefine,wang2023voyager,yang2024opro}. Recent surveys summarize this loop as generation, execution, evaluation, reflection, memory update, and sometimes model or prompt optimization \citep{gao2025selfevolvingagents,ren2026selfimprovingagents, ding2026memcompiler}. 

Embodied agents face a harder version of this problem because they operate in a continuous world: the harness must evaluate both robot state and world state, decide when to call policies, tools, critics, or recovery modules, and prevent early errors from becoming hard-to-recover physical failures. Moreover, current VLAs are far weaker and less reliable than frontier LLMs in their native digital domains, so embodied agents need more external tools, verifiers, memories, and frequent critics to support robust execution. Real-world policy-improvement systems such as ENPIRE, Visual Verification/VERITAS, Learning While Deploying, and SOP show that robot deployment can be organized as a closed loop of execution, verification, data selection, and policy update rather than a one-time training pipeline \citep{xiao2026enpire,li2025veritas,agibot2026lwd,agibot2026sop}. Skill-centric systems such as ASPIRE, ReSYNC, VASO, and EmbodiSkill instead emphasize discovering reusable programs, concepts, contracts, or skills from failures and reflections \citep{aspire2026,zhang2026resync,vaso2026formal,embodiskill2026}. Test-time and reasoning-oriented systems such as RoboTTT, R\&B-EnCoRe, EEAgent, and SEEA-R1 further show that adaptation can happen through long context, richer feedback, reflection, or reinforcement tuning \citep{robottt2026,renbencore2026,wang2026eeagent,seea2025,ding2025adanav}. 

Existing systems usually improve one layer of the stack, such as code, prompts, verifiers, memories, or policy data, but few jointly address high-frequency runtime critics, recoverable failure traces, reusable recovery skills, and high-throughput rollout execution. Our design targets this combined gap by grounding reflection in runtime critics and recovery actions, then using rollout infrastructure to repeatedly evaluate, replay, and improve embodied agents.

\subsection{Rollout Infrastructure Gaps for Embodied Self-Evolution}
There is still little infrastructure work designed specifically for self-evolving embodied agents. Most mature rollout systems come from general RL or distributed learning, such as A3C-style asynchronous actors, IMPALA, SEED RL, RLlib/Ray, Acme, and recent large-scale RL systems such as RLinf \citep{mnih2016a3c,espeholt2018impala,espeholt2020seed,liang2018rllib,hoffman2020acme,yu2025rlinf}. These systems established important ideas such as actor-learner separation, scalable experience collection, centralized inference, and flexible execution graphs. 

However, \textbf{embodied rollouts add workloads that are not central in standard rollout runtimes}: VLA inference, diffusion or flow action heads, perception encoders, tool calls, memory retrieval, simulators, real robot workers, resets, safety checks, and verifier/critic models. As a result, throughput is limited not only by environment stepping, but also by heterogeneous model serving and coordination across CPU, GPU, robot, and cloud resources. Another bottleneck is extensibility: adding a new simulator, tool, model backend, or agent variant often requires changing orchestration code instead of registering a new component behind a stable interface. These limits directly slow self-evolution because fewer failures can be discovered, diagnosed, replayed, and converted into reusable improvements per unit time. Our rollout infrastructure is designed for this embodied workload mix through decoupled execution, continuous batching, resource-sharing environments, and fine-grained processor control.

%% file: tex/conclusion.tex
\section{Conclusion}
We presented \sys, a closed-loop embodied harness that closes the gap between static, open-loop agent harnesses and the high-frequency governance that physical execution demands, by evolving code-based runtime critics and recovery skills online while keeping the base policy model frozen. Through three coordinated loops operating at action, rollout-batch, and iteration timescales, \sys turns accumulated rollout experience into validated, generalizable improvements in execution behavior. To sustain this evolution, we built Z-Infra, the first rollout infrastructure designed for self-evolving embodied agents, decoupling agent logic from heterogeneous execution resources. Our experiments on LIBERO and RoboCasa show that this self-evolution scales task success toward the frozen policy's capability ceiling and transfers across tasks, demonstrating that harness self-evolution is a viable path toward reliable embodied intelligence.

Looking ahead, we plan to extend \sys and Z-Infra to real robots: enabling fast, massively parallel rollout collection and self-evolution directly for real machines by bridging the sim-to-real gap, and integrating real-robot environments as first-class workers within Z-Infra alongside simulated ones.

%% file: tex/appendix.tex
\section{RoboCasa Atomic Task Mapping}
\label{app:atomic-task-mapping}

Table~\ref{tab:atomic-task-mapping} maps the compact identifiers used in
Table~\ref{tab:robocasa-atomic} to the official task names in the
\href{https://robocasa.ai/docs/build/html/tasks/atomic_tasks.html}{RoboCasa
1.0.1 Atomic Tasks documentation}. These 18 tasks constitute the complete
Atomic-Seen split.

\begin{table}[t]
    \centering
    \caption{Mapping from compact identifiers to official RoboCasa task names.}
    \label{tab:atomic-task-mapping}
    \small
    \setlength{\tabcolsep}{3pt}
    \renewcommand{\arraystretch}{1.08}
    \begin{minipage}[t]{0.49\linewidth}
        \centering
        \begin{tabular}{cl}
            \toprule
            \textbf{ID} & \textbf{Official task name} \\
            \midrule
            T1 & \texttt{NavigateKitchen} \\
            T2 & \texttt{TurnOnMicrowave} \\
            T3 & \texttt{PickPlaceCounterToStove} \\
            T4 & \texttt{PickPlaceSinkToCounter} \\
            T5 & \texttt{PickPlaceDrawerToCounter} \\
            T6 & \texttt{PickPlaceCounterToCabinet} \\
            T7 & \texttt{PickPlaceToasterToCounter} \\
            T8 & \texttt{TurnOnSinkFaucet} \\
            T9 & \texttt{CoffeeSetupMug} \\
            \bottomrule
        \end{tabular}
    \end{minipage}\hfill
    \begin{minipage}[t]{0.49\linewidth}
        \centering
        \begin{tabular}{cl}
            \toprule
            \textbf{ID} & \textbf{Official task name} \\
            \midrule
            T10 & \texttt{OpenCabinet} \\
            T11 & \texttt{CloseFridge} \\
            T12 & \texttt{SlideDishwasherRack} \\
            T13 & \texttt{TurnOnElectricKettle} \\
            T14 & \texttt{OpenStandMixerHead} \\
            T15 & \texttt{CloseBlenderLid} \\
            T16 & \texttt{OpenDrawer} \\
            T17 & \texttt{CloseToasterOvenDoor} \\
            T18 & \texttt{TurnOffStove} \\
            \bottomrule
        \end{tabular}
    \end{minipage}
\end{table}

\section{LIBERO-Pro Task Mapping}
\label{app:libero-pro-task-mapping}

LIBERO-Pro extends the original LIBERO benchmark with controlled
perturbations for evaluating generalization under distribution shifts.
In the main text, we evaluate the LIBERO-Goal and LIBERO-10 suites, where
LIBERO-10 is denoted as ``Long'' for brevity. Each suite contains 10 tasks,
indexed as Task 0--9.

We consider two LIBERO-Pro perturbation settings. ``T'' denotes the
task/instruction-redirection perturbation, where the instruction is redirected
to another valid target object or goal condition. ``S'' denotes the
swap/position-swap perturbation, where the initial positions of relevant
objects are swapped or rearranged while the instruction remains fixed.

Table~\ref{tab:libero-pro-task-mapping} maps the compact task identifiers
used in the main text to their underlying LIBERO task instructions. For the
T setting, the table identifies the underlying task index; the evaluated
instruction can differ because of instruction redirection.

\begin{table*}[t]
    \centering
    \caption{\textbf{Mapping of LIBERO-Pro task identifiers used in the main
    text to the underlying LIBERO tasks.}
    ``Goal'' corresponds to LIBERO-Goal and ``Long'' corresponds to
    LIBERO-10. The same task indices are used for both T and S perturbation
    settings.}
    \label{tab:libero-pro-task-mapping}
    \small
    \setlength{\tabcolsep}{4pt}
    \renewcommand{\arraystretch}{1.08}
    \begin{tabular}{c p{0.36\textwidth} c p{0.46\textwidth}}
        \toprule
        \textbf{ID} & \textbf{LIBERO-Goal}
        & \textbf{ID} & \textbf{LIBERO-10 (Long)} \\
        \midrule
        Task 0 & open the middle drawer of the cabinet
        & Task 0 & put both the alphabet soup and the tomato sauce in the basket \\

        Task 1 & put the bowl on the stove
        & Task 1 & put both the cream cheese box and the butter in the basket \\

        Task 2 & put the wine bottle on top of the cabinet
        & Task 2 & turn on the stove and put the moka pot on it \\

        Task 3 & open the top drawer and put the bowl inside
        & Task 3 & put the black bowl in the bottom drawer of the cabinet and close it \\

        Task 4 & put the bowl on top of the cabinet
        & Task 4 & put the white mug on the left plate and put the yellow and white mug on the right plate \\

        Task 5 & push the plate to the front of the stove
        & Task 5 & pick up the book and place it in the back compartment of the caddy \\

        Task 6 & put the cream cheese in the bowl
        & Task 6 & put the white mug on the plate and put the chocolate pudding to the right of the plate \\

        Task 7 & turn on the stove
        & Task 7 & put both the alphabet soup and the cream cheese box in the basket \\

        Task 8 & put the bowl on the plate
        & Task 8 & put both moka pots on the stove \\

        Task 9 & put the wine bottle on the rack
        & Task 9 & put the yellow and white mug in the microwave and close it \\
        \bottomrule
    \end{tabular}
\end{table*}

\section{Case Study: Critic-Guided Recovery for PnP-Stove}
\label{app:pnp-stove-skill}

We use \texttt{PickPlaceCounterToStove} (PnP-Stove) as a representative
example of an evolved skill.  The VLA remains frozen and executes the nominal
task policy.  During execution, a lightweight critic monitors task progress
and physical grasp state, including the official grasp predicate, bilateral
finger contact, object--gripper drift, transport progress, placement, and
post-release separation.  The critic only produces a structured proposal; it
does not execute an action or declare success.  The Orchestrator decides
whether to continue the VLA or invoke a recovery.

The evolution adds three reusable
capabilities.  First, an object-relative pregrasp moves the gripper to a more
reliable acquisition pose.  Second, grasp failure triggers bounded regrasp:
the robot releases, restages, and selects a new proposal that is different
from previously failed grasps.  Third, placement recovery lowers the object
until support is detected, releases it, and retreats until the gripper is
safely separated.  Algorithm~\ref{alg:pnp-stove-simple} summarizes the complete
critic--recovery loop.

\begin{algorithm}[H]
\DontPrintSemicolon
\caption{Simplified critic-guided skill for PnP-Stove}
\label{alg:pnp-stove-simple}
\KwIn{Frozen VLA $\pi$; runtime critic $C$; recovery library $R$;
failed-grasp memory $\mathcal{M}$}
\KwOut{Official task success or a bounded failure record}

\While{the episode is active}{
    execute one VLA or recovery action chunk and collect state window $W_t$\;
    $P_t\leftarrow C(W_t)$\tcp*[r]{proposal only}

    \If{the official task predicate is satisfied}{
        \KwRet{success}\;
    }

    \uIf{$P_t$ reports normal progress}{
        continue the frozen VLA\;
    }
    \uElseIf{$P_t$ reports failed or unstable grasp}{
        freeze unsafe motion, release, and restage\;
        obtain a fresh object-relative grasp not equivalent to $\mathcal{M}$\;
        execute pregrasp, acquisition, and a short stability check\;
        if the retry fails, add it to $\mathcal{M}$ and repeat within budget\;
    }
    \uElseIf{$P_t$ reports stalled transport with a stable grasp}{
        preserve the grasp and switch from coarse base carry to fine arm alignment\;
    }
    \uElseIf{$P_t$ reports supported placement}{
        release the object and retreat until the gripper-far predicate holds\;
    }

    the Orchestrator approves every intervention and any return to nominal execution\;
}
\KwRet{bounded failure record}
\end{algorithm}

This skill transfers because its critics and recoveries are defined by
object-relative geometry and generic physical predicates rather than a
stove-specific image template or a replayed source trajectory.  Related
pick-and-place tasks can therefore reuse the same grasp, transport, and
placement recovery logic by rebinding the live object and target receptacle.

\section{Case Study: Critic-Guided Recovery for Libero-Pro Goal-T2}
\label{app:libero-goal-t2-skill}

We use \texttt{Goal-T2: PutWineBottleInBowl} as a representative
example of an evolved Libero-Pro skill.  The VLA remains frozen and executes
the nominal task policy.  During execution, a lightweight critic monitors
task progress and physical interaction state, including the requested action,
realized end-effector motion, gripper aperture, finger--object contact,
object--gripper drift, grasp retention, target-relative transport, and the
official task predicate.  The critic only produces a structured proposal; it
does not execute an action or declare success.  The Orchestrator decides
whether to continue the VLA or invoke a recovery.

The evolution adds three reusable capabilities.  First, an object-relative
pregrasp recovery moves the gripper to a collision-safe pose before
re-attempting acquisition.  Second, a retained-object critic distinguishes
successful transport from motion with an empty or slipping gripper; grasp loss
triggers bounded release, restaging, and regrasp from a different proposal.
Third, placement recovery aligns the retained bottle with the live bowl,
lowers it until containment is established, releases it, and retreats without
disturbing the placed object.  Algorithm~\ref{alg:libero-goal-t2-simple}
summarizes the complete critic--recovery loop.

\begin{algorithm}[H]
\DontPrintSemicolon
\caption{Simplified critic-guided skill for Libero-Pro Goal-T2}
\label{alg:libero-goal-t2-simple}
\KwIn{Frozen VLA $\pi$; runtime critic $C$; recovery library $R$;
failed-grasp memory $\mathcal{M}$}
\KwOut{Official task success or a bounded failure record}

\While{the episode is active}{
    execute one VLA or recovery action chunk and collect state window $W_t$\;
    $P_t\leftarrow C(W_t)$\tcp*[r]{proposal only}

    \If{the official task predicate is satisfied}{
        \KwRet{success}\;
    }

    \uIf{$P_t$ reports normal progress}{
        continue the frozen VLA\;
    }
    \uElseIf{$P_t$ reports non-approach or failed acquisition}{
        freeze unsafe motion, open the gripper, and restage\;
        obtain a fresh object-relative pregrasp not equivalent to $\mathcal{M}$\;
        execute approach, acquisition, and a short lift check\;
        if the retry fails, add it to $\mathcal{M}$ and repeat within budget\;
    }
    \uElseIf{$P_t$ reports grasp loss or unstable retention}{
        stop transport before the empty gripper reaches the target\;
        release residual contact, return to pregrasp, and reacquire the bottle\;
    }
    \uElseIf{$P_t$ reports stable grasp near the bowl}{
        align the bottle with the live receptacle and lower it into containment\;
        release and retreat until gripper--object separation is confirmed\;
    }

    the Orchestrator approves every intervention and any return to nominal execution\;
}
\KwRet{bounded failure record}
\end{algorithm}

This skill transfers because its critics and recoveries are defined by
object-relative geometry, realized robot motion, and generic contact and
grasp-retention predicates rather than a Goal-T2-specific image template or a
replayed source trajectory.  Related Libero-Pro pick-and-place tasks can
therefore reuse the same pregrasp, retention, and placement recovery logic by
rebinding the live object and target receptacle.